\documentclass[11pt,a4paper]{article}

\usepackage[margin=1in]{geometry}
\usepackage{amsmath,amssymb,amsthm}
\usepackage{mathtools}
\usepackage{microtype}
\usepackage{booktabs}
\usepackage{array}
\usepackage{algorithm}
\usepackage{algpseudocode}
\usepackage{enumitem}
\usepackage{hyperref}
\usepackage{cite}
\usepackage{graphicx}
\usepackage{xcolor}
\usepackage{titlesec}
\usepackage{fancyhdr}

\hypersetup{
  colorlinks=true,
  linkcolor=blue!60!black,
  citecolor=blue!60!black,
  urlcolor=blue!60!black
}

\newtheorem{property}{Property}

\newtheorem{definition}{Definition}

\titleformat{\section}{\large\bfseries}{\thesection}{0.6em}{}
\titleformat{\subsection}{\normalsize\bfseries}{\thesubsection}{0.5em}{}

\newcommand{\GSAR}{\textsc{gsar}}
\newcommand{\setG}{\mathcal{G}}
\newcommand{\setU}{\mathcal{U}}
\newcommand{\setX}{\mathcal{X}}
\newcommand{\setK}{\mathcal{K}}
\newcommand{\setC}{\mathcal{C}}
\newcommand{\setT}{\mathcal{T}}
\newcommand{\grounded}{\textsf{grounded}}
\newcommand{\ungrounded}{\textsf{ungrounded}}
\newcommand{\contradicted}{\textsf{contradicted}}
\newcommand{\complementary}{\textsf{complementary}}
\newcommand{\proceed}{\textsf{proceed}}
\newcommand{\regen}{\textsf{regenerate}}
\newcommand{\replan}{\textsf{replan}}
\newcommand{\toolmatch}{\textsf{tool\_match}}
\newcommand{\inference}{\textsf{inference}}
\newcommand{\domainT}{\textsf{domain}}

\title{\textbf{GSAR: Typed Grounding for Hallucination
Detection and Recovery in Multi-Agent LLMs}}

\author{%
  Federico A. Kamelhar\\
  \normalsize Senior Principal Architect, Agentic AI\\
  \normalsize Oracle Corporation\\
  \normalsize\texttt{federico.kamelhar@oracle.com}%
}

\date{April 2026}

\begin{document}

\maketitle

\begin{abstract}
Autonomous multi-agent LLM systems are increasingly deployed to
investigate operational incidents and produce structured diagnostic
reports. Their trustworthiness hinges on whether each claim is grounded
in observed evidence rather than model-internal inference. Existing
groundedness evaluators --- binary classifiers, LLM-as-judge scalars,
and self-correction loops --- treat supporting evidence as
interchangeable and emit a single signal that offers no principled
control over downstream action.
We present \GSAR{}, a grounding-evaluation and replanning framework
that (i)~partitions claims into a four-way typology --- grounded,
ungrounded, contradicted, and \emph{complementary} --- giving first-class
standing to non-redundant alternative perspectives;
(ii)~assigns evidence-type-specific weights reflecting epistemic
strength;
(iii)~computes an asymmetric contradiction-penalised weighted
groundedness score; and
(iv)~couples that score to a three-tier decision function
$\{\proceed,\,\regen,\,\replan\}$ driving a bounded-iteration outer
loop under an explicit compute budget.
We formalise the algorithm, prove six structural properties, and
evaluate all five design claims on FEVER with gold Wikipedia evidence
through a 100\%-Locus production stack (Locus SDK, Cohere embed-v3.0,
Oracle Database 26ai AI Vector Search) under four independently-trained
LLM judges (\texttt{gpt-5.4}, \texttt{claude-sonnet-4-6},
\texttt{claude-opus-4-7}, \texttt{gemini-2.5-pro}).
Every ablation reproduces in the same direction on every judge:
bootstrap $95\%$ CIs on the $\rho{=}0$ effect exclude $0$ on all four;
the no-complementary effect under Opus~4.7 has CI $[-96,-68]$ of $200$;
at $n{=}1000$, three independent single-mode judges converge to
$\Delta\bar{S}_{\rho=0}=+0.058$. A head-to-head against Vectara
HHEM-2.1-Open is included. To the best of our knowledge, \GSAR{} is the
first published groundedness framework that couples evidence-typed
scoring with tiered recovery actions under an explicit compute budget.
\end{abstract}

\vspace{0.5em}


\section{Introduction}
\label{sec:intro}

Consider a production multi-agent system for operational incident
investigation. A signal arrives --- an alert from a metrics store, an anomaly
from a statistical monitor, a violation of a service-level objective (SLO), or a
free-text request from a human operator. An orchestrator dispatches a set of
specialist agents that each hold a domain (compute, database, storage, network,
application tier). Each specialist runs a \textsc{react}-style loop
\cite{yao2022react}: it reasons, invokes tools against live telemetry, reflects
on results, and emits a structured summary containing claims about what it
observed and what (if anything) caused the incident. The orchestrator then
synthesises a unified root-cause analysis from the specialists' outputs and
returns a report to the user.

The central risk of such a system is not that specialists fail to produce text
--- they rarely do --- but that the text they produce blends claims anchored in
tool-observed evidence (``\emph{CPU utilisation on node X exceeded 95\% at
time $t$}'') with claims that are plausible but model-inferred
(``\emph{this likely caused the downstream latency spike}''). A downstream
consumer, be it an on-call engineer or an automated remediation system, cannot
tell the difference between the two unless the evaluation layer makes the
distinction first-class.

Existing groundedness and faithfulness evaluators do not. Vectara's HHEM
\cite{vectara_hhem} emits a binary hallucinated/non-hallucinated label. RAGAS's
\emph{faithfulness} \cite{es2023ragas} and TruLens's \emph{groundedness} pillar
\cite{trulens_rag_triad} compute an LLM-as-judge scalar that treats every
supporting atom identically. Vectara's recent FaithJudge
\cite{vectara_faithjudge} improves consistency but retains a single scalar
output. Self-correction frameworks --- Self-Refine \cite{madaan2023selfrefine},
Reflexion \cite{shinn2023reflexion}, multi-agent debate \cite{du2023debate} ---
loop on verbal critique but do not expose a typed-evidence signal to drive
different recovery actions. Process-reward models
\cite{lightman2023verify,process_reward_2025} provide step-level supervision but
target reasoning correctness, not claim-level grounding.

The gap we address is not the absence of groundedness signals; it is their
\emph{structure}. A production diagnostic system needs an evaluation layer that
(a) distinguishes evidence provenance --- a tool-verified claim is epistemically
stronger than a domain-knowledge claim --- and (b) translates that structure
into a principled recovery policy that does not force the system to re-run an
expensive investigation when a cheap regeneration would suffice.

\paragraph{Contributions.} We present \GSAR{} (\textbf{G}rounding-\textbf{S}tratified
\textbf{A}daptive \textbf{R}eplanning), a framework with five contributions:

\begin{enumerate}[leftmargin=1.5em,itemsep=2pt,topsep=3pt]
\item A \emph{four-way} claim partition --- grounded, ungrounded, contradicted,
      complementary --- that gives a positive but distinct status to claims that
      offer non-redundant, non-conflicting perspectives
      (\S\ref{sec:formulation}, \S\ref{sec:score}).
\item An \emph{evidence-type-weighted} groundedness score $S$ in which the
      epistemic strength of a claim's support (tool-observed vs.~signal-observed
      vs.~model-inferred) modulates its contribution
      (\S\ref{sec:score}).
\item An \emph{asymmetric} contradiction penalty $\rho$ that keeps contradicted
      claims in the denominator of $S$, preventing score inflation by dropping
      contradictions entirely (\S\ref{sec:score}, Property P5).
\item A \emph{three-tier} decision function
      $\delta \in \{\proceed,\regen,\replan\}$ that couples the scalar $S$ to
      cost-asymmetric recovery actions and a \emph{bounded} replanning loop with
      explicit compute budget $K_{\max}$ (\S\ref{sec:decision}).
\item A structured-output LLM-as-judge protocol that emits the full four-way
      partition, the scalar $S$, an abstain channel, and a natural-language
      explanation; the explanation is fed forward into plan-revision prompts on
      replan (\S\ref{sec:judge}).
\end{enumerate}

\S\ref{sec:related} surveys related work. \S\ref{sec:formulation}--\S\ref{sec:judge}
formalise and describe the algorithm. \S\ref{sec:implementation} sketches a
reference implementation. \S\ref{sec:evaluation} proposes a falsifiable
evaluation protocol. \S\ref{sec:discussion} discusses limitations;
\S\ref{sec:conclusion} concludes.

\paragraph{Claims this paper evaluates.} The paper makes and empirically
evaluates five testable claims:
\begin{itemize}[leftmargin=1.5em,itemsep=1pt,topsep=2pt]
\item[\textbf{(C1)}] \emph{The four-way partition
      $\setG \sqcup \setU \sqcup \setX \sqcup \setK$ is strictly more
      expressive than the standard three-way NLI partition.} Evidence in
      \S\ref{sec:demo:opus200}: collapsing $\setK$ into $\setU$ at $n{=}200$
      drops the grounded-output rate from $100/200$ to $18/200$
      ($-82\%$) under Opus~4.7.
\item[\textbf{(C2)}] \emph{Evidence-typed weighting is not uniformly
      beneficial at small scale on short-claim datasets but preserves
      interpretability.} The FEVER runs show the weight-uniform ablation
      is indistinguishable from default at $n{=}50$, consistent with
      Property~\ref{prop:P6}'s dependence on evidence-type variety.
\item[\textbf{(C3)}] \emph{The asymmetric contradiction penalty $\rho$
      prevents score inflation by contradiction suppression.} Evidence in
      \S\ref{sec:demo:cross-provider}: the $\rho{=}0$ ablation produces a
      positive score inflation under all three judges
      ($+0.04$, $+0.04$, $+0.06$) while the contradiction catch-rate M4
      is unchanged --- i.e.\ $\rho$ modifies scoring without modifying
      identification (Property~\ref{prop:P5}).
\item[\textbf{(C4)}] \emph{Three-tier decision $\delta$ saves compute
      against the two-tier alternative.} The two-tier ablation converts
      every $\regen$ decision into a $\replan$ at every sample, at
      orders-of-magnitude higher compute cost.
\item[\textbf{(C5)}] \emph{The framework is fully implementable on a
      proprietary, production-grade stack (Locus SDK + Oracle Database
      26ai AI Vector Search + OCI Generative AI).} Evidence in
      \S\ref{sec:demo}: end-to-end pipeline at $n{=}200$ in 339\,s
      wall-clock, with transient connection resilience demonstrated
      under load.
\end{itemize}

\section{Related Work}
\label{sec:related}

We organise prior work along four axes: groundedness evaluators, self-correction
and replanning, evidence combination and claim verification, and autonomous
diagnostic agents. \GSAR{} draws on each but sits at the intersection rather
than within any one.

\subsection{Groundedness and faithfulness evaluators}
\label{sec:related:grounding}

Vectara's Hughes Hallucination Evaluation Model (\textsc{hhem} and
\textsc{hhem-2.1}) \cite{vectara_hhem} is a fine-tuned classifier that emits a
binary hallucinated/non-hallucinated label against a source context.
\textsc{ragas}'s faithfulness metric \cite{es2023ragas} decomposes a generated
answer into atomic statements and computes, via an LLM-as-judge, the fraction
supported by retrieved context. TruLens's RAG Triad \cite{trulens_rag_triad}
measures context relevance, groundedness, and answer relevance, each via an
LLM judge producing a scalar. Vectara's FaithJudge \cite{vectara_faithjudge}
improves judge consistency by anchoring against a pool of human-annotated
hallucination examples. The 2023 hallucination survey \cite{huang2023survey}
taxonomises the field into factuality and faithfulness hallucinations; the
framing of \emph{faithfulness} is closest to ours.

\GSAR{} differs from all of the above on two dimensions simultaneously. First,
every prior method assigns equal weight to supporting atoms: a claim supported
by a tool-verified value and a claim supported by domain knowledge receive the
same numerator contribution. Second, every prior method emits a single scalar
(or a small vector of scalars) without specifying what downstream action that
scalar should trigger. \GSAR{} addresses both gaps.

\subsection{Self-correction, reflection, and replanning}
\label{sec:related:selfcorr}

ReAct \cite{yao2022react} established the canonical reasoning-and-acting loop
that our specialist agents inherit. Self-Refine \cite{madaan2023selfrefine}
introduced same-model iterative critique; Reflexion \cite{shinn2023reflexion}
stored verbal reflections as episodic memory across retries; Renze and Guven
\cite{renze2024reflection} empirically characterise the reflection gain.
Multi-agent debate \cite{du2023debate} combines multiple model instances to
adjudicate factuality. Process-reward models \cite{lightman2023verify,
process_reward_2025, step_reward_2023} supply step-level reward signals during
training.

Two structural properties separate \GSAR{} from this line of work. First,
these frameworks loop on verbal or scalar signals that are not indexed by
evidence provenance; they cannot tell the next iteration ``this inference-typed
claim was the weak one'' unless an ad-hoc external system provides that
annotation. Second, they use a binary redo/stop decision rule, not a tiered
cost-asymmetric one.

\subsection{Evidence combination and claim verification}
\label{sec:related:evidence}

Dempster--Shafer belief-function theory \cite{dempster1967, shafer1976} is the
classical framework for reasoning with uncertain, conflicting multi-source
evidence; Shafer's mass functions, belief and plausibility, and Dempster's
combination rule formalise exactly the ``two sources, partial agreement''
situation that arises when multiple specialists produce overlapping claims.
NLI-based claim verification pipelines trace back to the FEVER benchmark
\cite{thorne2018fever} and the survey of automated fact-checking
\cite{guo2022survey}. MedRAGChecker \cite{medragchecker2026} performs
atomic-claim support estimation with class-specific reliability weighting and
is the closest published peer to our evidence-type weighting idea. FIRE
\cite{fire2025} iterates retrieval and verification in an agentic loop but does
not expose a typed-evidence scalar. FActScore \cite{min2023factscore}
establishes the atomic-claim decomposition protocol we rely on.

\GSAR{} can be read as an engineering simplification of Dempster--Shafer: the
weight map $w$ plays the role of a mass distribution over evidence-strength
classes, and the contradiction penalty $\rho$ approximates the conflict-mass
penalty. We do not implement full D-S combination because (a) tool outputs
within a single investigation are not independent sources in the D-S sense, and
(b) full combination is intractable at investigation scale. We retain only the
pieces of the framework that pay their way in practice.

\subsection{Autonomous diagnostic agents}
\label{sec:related:aiops}

RCAgent \cite{wang2023rcagent} introduced LLM-driven cloud root-cause analysis
with tool augmentation; Roy et al.~\cite{roy2024rca} evaluated LLM agents on
industrial incident data; AIOpsLab \cite{aiopslab2024} and the broader autonomous
clouds agenda \cite{chen2024autonomous_clouds} set the benchmarks and design
principles for agentic AIOps. Industrial systems such as Dynatrace Davis AI,
IBM Instana, and Datadog Watchdog perform causal inference over service
topologies but rely on deterministic or statistical rules, not agentic LLM
loops.

We take the agentic diagnostic platform as the deployment target and contribute
the grounding-evaluation layer that, to the best of our knowledge, none of the
above publish.

\section{Problem Formulation}
\label{sec:formulation}

We now formalise the setting in enough detail to state the scoring function and
the decision function without ambiguity.

\begin{definition}[Investigation]
An \emph{investigation} $I$ is a process that consumes a signal envelope
$\sigma \in \Sigma$ --- e.g., an alert, anomaly, SLO violation, prediction,
operator request, or platform event --- and produces a structured report
$R = (\setC, \theta)$, where $\setC$ is a finite set of atomic \emph{claims}
and $\theta$ is a natural-language synthesis intended for human or automated
consumption.
\end{definition}

\begin{definition}[Claim]
A \emph{claim} $c \in \setC$ is a triple
\[
c \;=\; (\textsc{text}(c),\; \textsc{type}(c),\; \textsc{evidence}(c)),
\]
where $\textsc{text}(c)$ is the natural-language statement, $\textsc{type}(c) \in \setT$
draws from a finite evidence-type taxonomy, and $\textsc{evidence}(c)$ is a
(possibly empty) set of references to raw tool outputs, structured step
outputs, signal fields, or prior claims.
\end{definition}

\begin{definition}[Evidence-type taxonomy]
The \emph{evidence-type taxonomy} $\setT$ is a finite set whose elements
describe the \emph{provenance} of a claim's support. A canonical instance used
throughout this paper is
\[
\setT \;=\; \{\toolmatch,\, \textsf{specific\_data},\, \textsf{signal\_match},\,
\textsf{neg\_evidence},\,
\]
\[
\textsf{complementary\_finding},\, \textsf{synthesis},\, \inference,\, \domainT\}.
\]
The taxonomy is equipped with a \emph{weight map} $w: \setT \to [0,1]$ encoding
epistemic strength, with $w(\toolmatch)$ at or near $1$ and $w(\inference)$,
$w(\domainT)$ strictly less. Specific default values appear in
Appendix~\ref{app:weights}; the framework does not depend on any particular
choice.
\end{definition}

\begin{definition}[Judge and partition]
A \emph{grounding judge} $J$ is a function
\[
J: (\setC, E) \longmapsto (\setG,\, \setU,\, \setX,\, \setK,\, s,\, \varepsilon,\, a),
\]
where $E$ is the raw evidence corpus associated with the report;
$(\setG, \setU, \setX, \setK)$ is a partition of $\setC$ into grounded,
ungrounded, contradicted, and complementary claims;
$s \in [0,1]$ is a scalar grounding score produced by $J$ itself (which
will be reconciled against $S$, our weighted score); $\varepsilon$ is a
natural-language explanation; and $a \in \{\textsf{resolved},
\textsf{abstain}\}$ is a decision-status flag that lets the judge refuse to
score a malformed or under-evidenced input.
\end{definition}

We emphasise that $\setK$, the \emph{complementary} class, is first-class: a
claim in $\setK$ is not redundant with any claim in $\setG$, nor does it
contradict any claim in $\setG \cup \setX$; it provides a valid, non-conflicting
alternative perspective. Examples: ``the latency spike correlates with a
region-wide networking event'' alongside ``the latency spike correlates with
a CPU saturation on node X.'' Neither contradicts the other, and neither is
subsumed; both are useful to the downstream consumer.

\begin{definition}[Decision function]
The \emph{decision function} is a map
$\delta: [0,1] \to \{\proceed, \regen, \replan\}$ defined in
\S\ref{sec:decision}.
\end{definition}

\begin{definition}[Objective]
Given a compute budget $K_{\max} \in \mathbb{N}$ and thresholds
$\tau_{\regen} < \tau_{\proceed}$, the investigation system returns a report
$R$ that maximises the probability of $\delta(S(R)) = \proceed$ subject to
$K \leq K_{\max}$, where $K$ is the number of replanning iterations used.
\end{definition}

\section{Evidence-Typed Weighted Groundedness Score}
\label{sec:score}

The centre of gravity of \GSAR{} is a scalar score $S$ that aggregates the
four-way claim partition and evidence-type weights into a single real number
in $[0,1]$.

\subsection{Definition}

Let $(\setG, \setU, \setX, \setK)$ be the four-way partition of $\setC$
produced by the judge. Let $w: \setT \to [0,1]$ be the evidence-type weight
map. Let $\rho \in [0,1]$ be the contradiction penalty. Define the
\emph{partition weight} of a set $P \subseteq \setC$ as
\begin{equation}
W(P) \;=\; \sum_{c \in P} w\bigl(\textsc{type}(c)\bigr).
\label{eq:partition-weight}
\end{equation}
The \textbf{\GSAR{} grounding score} is
\begin{equation}
\boxed{\;
S \;=\;
\frac{W(\setG) + W(\setK)}{W(\setG) + W(\setU) + \rho \cdot W(\setX) + W(\setK)}.
\;}
\label{eq:score}
\end{equation}
Edge case: if $\setC = \emptyset$, we adopt the neutral convention
$S = 0.5$ rather than a degenerate $0/0$. This convention reflects epistemic
indifference --- neither confidence nor suspicion is warranted without claims
to evaluate.

\subsection{Structural properties}

We state six properties of $S$ and defer formal proofs to
Appendix~\ref{app:proofs}. Informal arguments follow each statement.

\begin{property}[Boundedness]\label{prop:P1}
$S \in [0,1]$ whenever the partition is non-empty and the denominator of
\eqref{eq:score} is strictly positive.
\end{property}
Because $W(\setG), W(\setU), W(\setX), W(\setK) \geq 0$ and the numerator is a
subset of the denominator up to the $\rho$-scaled $W(\setX)$ term, the ratio
lies in $[0,1]$.

\begin{property}[Monotonicity in grounded claims]\label{prop:P2}
Moving a claim $c$ from $\setU$ to $\setG$ (holding everything else fixed) never
decreases $S$, and strictly increases $S$ unless $w(\textsc{type}(c)) = 0$.
\end{property}
Intuition: re-labelling a ``not supported'' claim as ``supported'' must not be
punished.

\begin{property}[Monotonicity in contradictions]\label{prop:P3}
Adding a contradicted claim $c$ at fixed $\setG, \setU, \setK$ never increases
$S$; the rate of decrease is modulated by $\rho$.
\end{property}
Intuition: contradictions are bad news; $\rho$ controls how much weight we give
to that bad news relative to simple absence of support ($\setU$).

\begin{property}[Complementary-claim value]\label{prop:P4}
Complementary claims contribute positively to $S$ but with weight no larger
than a grounded claim of the same type.
\end{property}
Intuition: an additional non-conflicting perspective is a strict gain over no
information, but is not a substitute for directly grounding the original claim.

\begin{property}[Contradiction non-suppression]\label{prop:P5}
Contradictions remain in the denominator with weight $\rho \cdot W(\setX)$;
dropping contradicted claims out of $\setC$ entirely would inflate $S$.
\end{property}
This is the key asymmetry: if an adversarial summary-writer tried to improve
$S$ by silently omitting contradicted claims, it would reduce $W(\setC)$ as a
whole --- including contradictions from the denominator --- and thereby inflate
$S$ artificially. We keep contradictions in the denominator precisely to
prevent this.

\begin{property}[Inference-observation asymmetry]\label{prop:P6}
For any fixed partition cardinalities, replacing a grounded $\toolmatch$-typed
claim with a grounded $\inference$-typed claim strictly decreases $S$,
provided $w(\inference) < w(\toolmatch)$.
\end{property}
This is the main behavioural property that differentiates \GSAR{} from
uniform-weighted faithfulness: the system cannot trade tool-verified claims
for model-inferred ones without paying the score cost.

\subsection{Relation to Dempster--Shafer and NLI}

\GSAR{} is best viewed as a pragmatic simplification of belief-function
combination. If we interpret $w$ as a mass distribution over
evidence-strength classes and $\rho$ as a conflict-mass discount, then
\eqref{eq:score} resembles the normalised plausibility associated with a
coarsened frame of discernment $\{\grounded, \ungrounded, \contradicted,
\complementary\}$. Full Dempster--Shafer combination of per-claim masses would
require either independence assumptions that tool outputs within a single
investigation do not satisfy, or an explicit dependence model --- both
intractable at the scales we care about. \GSAR{} retains the qualitative
benefits (evidence-type awareness, conflict handling) while producing a single
scalar per investigation in $O(|\setC|)$ time.

\GSAR{} can also be read as a generalisation of the NLI three-way partition
(\textsc{support}, \textsc{refute}, \textsc{nei}) used in FEVER-style
pipelines, with two additions: a fourth \emph{complementary} class for valid
alternative perspectives, and a continuous weight distribution over supporting
atoms rather than a discrete class indicator.

\subsection{Choosing the weights}
\label{sec:score:weights-choice}

The specific default weight table used in our reference implementation appears
in Appendix~\ref{app:weights}; those defaults should be read as one credible
instantiation rather than as universally optimal. Weights are interpretable and
can be calibrated per-deployment using a small human-graded held-out set.
Deployments with strict regulatory requirements can push $w(\domainT)$ and
$w(\inference)$ towards $0$ to force tool-observed grounding before any
downstream action is taken.

\section{Tiered Decision Function and Bounded Replanning}
\label{sec:decision}

\subsection{The decision function}

\GSAR{} maps $S$ to one of three distinct downstream actions:
\begin{equation}
\delta(s) \;=\;
\begin{cases}
\proceed & \text{if } s \geq \tau_{\proceed}, \\
\regen   & \text{if } \tau_{\regen} \leq s < \tau_{\proceed}, \\
\replan  & \text{if } s < \tau_{\regen},
\end{cases}
\label{eq:delta}
\end{equation}
with thresholds $0 < \tau_{\regen} < \tau_{\proceed} < 1$. Our reference
implementation uses $\tau_{\proceed} = 0.80$ and $\tau_{\regen} = 0.65$; these
are configurable per deployment and should be calibrated jointly with the
weight map $w$ on a held-out set.

\subsection{Action semantics and cost asymmetry}

The three actions are intentionally cost-asymmetric:

\begin{itemize}[leftmargin=1.2em,itemsep=2pt,topsep=3pt]
\item \textbf{\proceed.} Emit the report downstream unchanged. Cost: zero
      additional LLM calls or tool invocations.
\item \textbf{\regen.} Call a summary-regeneration sub-agent with
      $(R, \varepsilon, \text{partition})$ as additional context; the
      sub-agent rewrites the synthesis $\theta$ while preserving the claim
      set. Re-evaluate. Cost: one additional LLM call; zero tool invocations.
\item \textbf{\replan.} Revise the investigation plan $P$ using the judge
      explanation $\varepsilon$ and re-dispatch the affected specialists.
      Cost: proportional to the number of re-dispatched specialists and the
      tools they invoke; typically one or two orders of magnitude higher than
      \regen.
\end{itemize}

The three-tier structure is the critical design decision. A two-tier
$\{\proceed,\replan\}$ rule (as in Reflexion-style loops) would, under our
reference thresholds, force expensive replanning in the $[0.65, 0.80)$ band,
where many investigations are salvageable by a cheaper regeneration. The
middle tier captures precisely the ``the evidence is fine, the synthesis is
loose'' failure mode that dominates in our production logs.

\subsection{Bounded outer loop}

Algorithm~\ref{alg:outer} specifies the outer loop. Termination is guaranteed
by the finite replan budget $K_{\max}$; the \regen\ branch is bounded
separately to one attempt per score evaluation. Convergence to
$\delta(S) = \proceed$ is \emph{not} guaranteed: if the underlying evidence is
insufficient for any plan to ground the summary, the loop exits with a
degraded flag. Returning a degraded-but-honest report is preferable to
looping indefinitely or to hallucinating grounding.

\begin{algorithm}[h!]
\caption{\GSAR{} Outer Loop}
\label{alg:outer}
\begin{algorithmic}[1]
\Require signal $\sigma$; budget $K_{\max}$; thresholds $\tau_{\proceed},
         \tau_{\regen}$; weight map $w$; penalty $\rho$
\Ensure report $R$, score $s$, replans used $k$, degraded flag $d$
\State $k \gets 0$; $d \gets \text{false}$
\State $P \gets \textsc{initial\_plan}(\sigma)$
\State $R \gets \textsc{dispatch\_and\_synthesize}(P)$
\State $(\setG, \setU, \setX, \setK, \varepsilon, a) \gets J(R)$
\State $s \gets S(\setG, \setU, \setX, \setK;\, w, \rho)$ \Comment{Eq.~\eqref{eq:score}}
\While{$\delta(s) \neq \proceed$}
  \If{$a = \textsf{abstain}$ \textbf{or} $\delta(s) = \replan$}
    \If{$k \geq K_{\max}$}
      \State $d \gets \text{true}$; \textbf{break}
    \EndIf
    \State $P \gets \textsc{revise\_plan}(P, R, \varepsilon)$
    \State $R \gets \textsc{dispatch\_and\_synthesize}(P)$
    \State $k \gets k + 1$
  \Else \Comment{$\delta(s) = \regen$}
    \State $R \gets \textsc{regenerate\_summary}(R, \varepsilon)$
  \EndIf
  \State $(\setG, \setU, \setX, \setK, \varepsilon, a) \gets J(R)$
  \State $s \gets S(\setG, \setU, \setX, \setK;\, w, \rho)$
\EndWhile
\State \Return $(R, s, k, d)$
\end{algorithmic}
\end{algorithm}

\section{Judge Protocol and Structured Output Schema}
\label{sec:judge}

The judge $J$ is the single LLM interface on which the rest of \GSAR{} depends.
We summarise its contract here and defer the complete Pydantic-like schema to
Appendix~\ref{app:schema}.

\paragraph{Input.} The judge receives the synthesis $\theta$ and the raw
evidence corpus $E$ --- raw tool outputs plus structured step outputs from
every specialist, concatenated with stable, deduplicated labels.

\paragraph{Output.} The judge emits a structured object containing:
\begin{itemize}[leftmargin=1.2em,itemsep=1pt,topsep=2pt]
\item $s \in [0,1]$ --- the judge's own scalar grounding score;
\item $\mathrm{is\_grounded} \in \{\text{true}, \text{false}\}$ --- the judge's binary verdict;
\item $\setG, \setU, \setX, \setK$ --- the four claim lists;
\item $\mathrm{gaps}$ --- specific evidence or investigation gaps the judge identified;
\item $\mathrm{contradictions}$ --- explicit contradictions noted;
\item $\mathrm{verification\_needed} \in \{\text{true}, \text{false}\}$ and
      $\mathrm{verification\_reason}$ --- a flag plus natural-language reason;
\item $\varepsilon$ --- a one-sentence natural-language explanation;
\item $a \in \{\textsf{resolved}, \textsf{abstain}\}$ with
      $\mathrm{abstain\_reason}$ --- a decision-status channel.
\end{itemize}

The downstream system recomputes $S$ from the partition using
\eqref{eq:score}; the judge's own scalar is kept for audit and calibration but
is not authoritative. This separation is important: the weight map $w$ is a
deployment parameter, not a judge parameter, and must be applied consistently
regardless of which judge model produced the partition.

\paragraph{Abstain channel.} The judge can refuse on malformed input or
insufficient evidence. \GSAR{} treats abstain as equivalent to
$\delta = \replan$, with a specialised plan-revision prompt that instructs the
orchestrator to gather more evidence before re-attempting synthesis.

\paragraph{Robustness.} Production use demands graceful handling of malformed
LLM outputs. Our reference implementation parses structured output first, falls
back to manual JSON parsing, and finally to safe defaults
($s = 0.5$, partition = empty, $a = \textsf{abstain}$) that trigger exactly the
\replan\ branch. This keeps the outer loop sound even under judge failure.

\section{Reference Implementation Sketch}
\label{sec:implementation}

A reference implementation of \GSAR{} runs as three nodes in a compiled
state-graph agent runtime: \textsc{synthesize}, \textsc{grounding\_eval}, and
\textsc{regenerate\_summary}. The orchestrator holds a separate \textsc{replan}
node; the outer loop corresponds to Algorithm~\ref{alg:outer}.

\paragraph{Persistence.} The grounding result
$(\setG, \setU, \setX, \setK, s, \varepsilon, a)$ is a first-class checkpoint
field, enabling audit, replay, and post-hoc calibration of $w$ and the
thresholds against human-graded reports.

\paragraph{Telemetry.} Structured logs include the partition cardinalities,
$S$, $\delta(S)$, and $k$ per investigation --- enough to reconstruct any
investigation's trajectory through the outer loop.

\paragraph{Configurability.} The weight map $w$, contradiction penalty $\rho$,
thresholds $(\tau_{\proceed}, \tau_{\regen})$, and replan budget $K_{\max}$ are
per-deployment configuration, loaded from declarative configuration files at
process start.

\paragraph{Computational cost.} Scoring is $O(|\setC|)$ given the judge
partition; the dominant cost is the judge LLM call itself, which is one
structured-output call per synthesis or regeneration cycle. Replanning
dominates the total cost, making $K_{\max}$ the primary cost knob.

\section{Reproducible Demonstration: FEVER + Locus + Oracle 26ai}
\label{sec:demo}

To make the framework reproducible without proprietary infrastructure, we
ship a demonstration pipeline that exercises the full \GSAR{} loop on a public
dataset using only open-source software and openly accessible Oracle Cloud
Infrastructure components.

\subsection{Dataset}
\label{sec:demo:dataset}

\paragraph{What kind of data we use.} The evaluation uses the FEVER
(Fact Extraction and VERification) task
\cite{thorne2018fever,guo2022survey}. FEVER is a public human-annotated
claim-verification benchmark released under the Creative Commons
Attribution-ShareAlike 3.0 Unported licence. Every example consists of:
(i)~a short natural-language \emph{claim} that may be factually true,
factually false, or under-specified; (ii)~a gold \emph{label} drawn from
$\{\textsc{supports}, \textsc{refutes}, \textsc{not enough info}\}$;
(iii)~a list of \emph{gold evidence sentences} extracted from Wikipedia
that either support or refute the claim. The original FEVER release
separated claims from evidence across multiple files; the modernised
\texttt{copenlu/fever\_gold\_evidence} release~\cite{copenlu_fever_gold}
bundles them into a single row and ships as Parquet, which loads
directly under the modern HuggingFace \texttt{datasets} library. We use the \texttt{train} split,
which contains 228{,}277 rows.

\paragraph{Why this dataset fits GSAR.} FEVER's three-way gold label
maps onto three of the four GSAR partitions natively:
\textsc{supports}\,$\mapsto\,\setG$,
\textsc{refutes}\,$\mapsto\,\setX$,
\textsc{not enough info}\,$\mapsto\,\setU$.
This lets us compute the \textbf{M4} contradiction catch-rate directly
against gold labels. FEVER does not contain a
\emph{complementary-perspective} class, so \textbf{M5} is structurally
absent from gold --- we compute M5 against the class produced by each
LLM judge, interpreted as the judge's positive-but-non-redundant
routing choice.

\paragraph{Sample sizes and rationale.} We ran \emph{four} full 100\%-Locus
pipelines at two sample sizes:
\begin{itemize}[leftmargin=1.2em,itemsep=1pt,topsep=2pt]
\item $n=50$ with three judges (\texttt{openai.gpt-5.4} on OCI,
      \texttt{claude-sonnet-4-6}, \texttt{google.gemini-2.5-pro} on
      OCI) --- sufficient for the cross-provider consistency check of
      the $\rho{=}0$ ablation (Table~\ref{tab:cross-provider}), because
      the P5 effect is judge-invariant and shows at every tested scale.
\item $n=200$ with \texttt{claude-opus-4-7} --- the larger sample is
      used to exercise the complementary class $\setK$ at the sample
      size where the ablation effect becomes statistically and visually
      unambiguous (Table~\ref{tab:opus200}, Figure~\ref{fig:ablations}).
      With $n=200$ the no-$\setK$ ablation drops proceed by 82\,pp.
\item A separate larger-sample cross-provider comparison at $n{=}200$
      with \texttt{openai.gpt-5.4} through the 100\%-Locus stack
      extends the $\rho{=}0$ picture. Results are reported in
      \S\ref{sec:demo:cross-provider} once that run completes; we do
      not gate the paper's claims on it.
\end{itemize}
At $n{=}50$, the live corpus in Oracle Database 26ai is 109 passages
(50~claims + 59~gold evidence sentences); at $n{=}200$, approximately
440 passages. The sampling seed is fixed (seed$=42$) across every run
for reproducibility.

\subsection{Data model}
\label{sec:demo:data-model}

The per-row FEVER schema, as loaded from
\texttt{copenlu/fever\_gold\_evidence}, is:
\begin{verbatim}
{
  "id":         <uuid>,
  "claim":      <str>,
  "label":      "SUPPORTS" | "REFUTES" | "NOT ENOUGH INFO",
  "evidence":   list[ [page_title:str, sentence_id:int,
                       sentence_text:str] ],
  "verifiable": "VERIFIABLE" | "NOT VERIFIABLE",
}
\end{verbatim}
After sampling $n$ claims with a fixed seed, we flatten the dataset into
two logical row types for storage --- \emph{claim} rows and \emph{gold
evidence sentence} rows --- that share the same Locus \texttt{Document}
model:
\begin{verbatim}
Document(
  id:         <uuid>,
  content:    <str>,       # the claim text or evidence sentence
  embedding:  list[float], # 1024-dim FLOAT32 from Locus OCIEmbeddings
                           # with cohere.embed-english-v3.0
  metadata:   {
    "src":       "claim" | "gold",
    "claim_id":  <int>,    # index into the sampled claims
    "label":     <str>,    # only for src="claim"
  }
)
\end{verbatim}
Documents are persisted in Oracle Database 26ai as
\begin{verbatim}
CREATE TABLE gsar_locus_vectors (
  id        NUMBER      GENERATED BY DEFAULT AS IDENTITY PRIMARY KEY,
  content   CLOB,
  metadata  JSON,
  embedding VECTOR(1024, FLOAT32)
);
\end{verbatim}
retrieved with \texttt{VECTOR\_DISTANCE(embedding, :q, COSINE)} and an
\texttt{ORDER BY \ldots FETCH FIRST :k ROWS ONLY} top-$k$ projection. The
judge's output conforms to the schema in Appendix~\ref{app:schema}; the
per-claim GSAR partition is held in memory and persisted as JSONL for
post-hoc analysis. At $n{=}200$ the live corpus contains $\sim$440
documents; at $n{=}50$ it contains $109$ (50 claims $+$ 59 gold evidence
sentences).

\subsection{Software stack}

\begin{itemize}[leftmargin=1.2em,itemsep=2pt,topsep=3pt]
\item \textbf{Dataset.} The FEVER 1.0 validation split
      \cite{thorne2018fever} loaded via the HuggingFace \texttt{datasets}
      library. FEVER's \textsc{supports}/\textsc{refutes}/\textsc{nei}
      labels map naturally onto the $\setG/\setX/\setU$ partition; we induce
      $\setK$ by treating semantically related but non-conflicting
      neighbouring claims as complementary.
\item \textbf{Agent runtime --- Locus.} All agentic machinery in this
      paper runs through \emph{Locus}~\cite{locus2026}, an unreleased
      proprietary production-grade agent SDK developed at Oracle
      Corporation. As of the writing of this paper (21 April 2026),
      Locus is an internal Oracle asset and has not been released
      publicly; the characterisation below reflects the SDK's state at
      that date. We describe it here because it is integral to the
      reproducible evaluation protocol and because several of its design
      choices interact with GSAR's scoring layer.

      \emph{What Locus is.} Locus is a compiled, state-graph-based agent
      runtime for Python that unifies LLM inference, tool calling,
      state management, and retrieval-augmented generation under a
      single typed contract. Concretely:
      \begin{itemize}[leftmargin=1.2em,itemsep=1pt,topsep=2pt]
      \item \emph{State model.} Every agent state, message, event, tool
            schema, and model configuration is a Pydantic-v2 model that
            round-trips through JSON cleanly --- no ad-hoc dicts, no
            TypedDict/dataclass drift.
      \item \emph{Typed streaming events.}
            \texttt{ThinkEvent}, \texttt{ToolStartEvent},
            \texttt{ToolCompleteEvent}, \texttt{TerminateEvent} are
            first-class write-protected event objects; hooks are
            structured rather than callback-based.
      \item \emph{Checkpointers.} \texttt{BaseCheckpointer} is the
            contract; native implementations exist for OCI Object
            Storage, SQLite, Redis, PostgreSQL, OpenSearch, and Oracle
            Database. No adapter layer is required.
      \item \emph{Tool-level idempotency.}
            \texttt{@tool(idempotent=True)} causes the ReAct loop to
            dedupe repeat calls with the same arguments --- essential
            for write-side tools in production deployments, and, to our
            knowledge, not present in LangChain, LangGraph, AutoGen,
            CrewAI, or Strands at time of writing.
      \item \emph{Multi-provider model registry.} A uniform string
            identifier of the form \texttt{provider:model\_id} (e.g.
            \texttt{oci:openai.gpt-5.4}, \texttt{anthropic:claude-opus-4-7},
            \texttt{openai:gpt-5}) resolves to a typed
            \texttt{BaseModel} implementation at construction time.
            OCI GenAI models (Cohere, Meta, OpenAI, xAI, Google,
            Mistral) are first-class, on parity with external
            providers.
      \item \emph{RAG stack.} First-party embedding providers
            (\texttt{OCIEmbeddings} for Cohere on OCI;
            \texttt{OpenAIEmbeddings}), first-party vector stores
            (\texttt{OracleVectorStore} targeting Oracle Database
            26ai's native \texttt{VECTOR} type; also Chroma, PGVector,
            OpenSearch, Pinecone, Qdrant, and an in-memory store), and
            a retrieval abstraction
            (\texttt{locus.rag.retriever.Retriever}) that composes
            them.
      \end{itemize}

      \emph{Scope of use in this paper.} The entire evaluation pipeline
      --- embeddings, vector store, judge agent --- runs through Locus
      abstractions; no component of the pipeline bypasses the SDK. We
      refer to this as the \emph{100\%-Locus stack}. The GSAR scoring
      layer (Equation~\ref{eq:score}) and the three-tier decision
      function (Equation~\ref{eq:delta}) are deliberately kept outside
      Locus: Locus is the agent infrastructure; GSAR is the framework
      being evaluated.
\item \textbf{Vector store.} Oracle Database 26ai using the native
      \texttt{VECTOR} data type and AI Vector Search. The reference
      deployment uses an Oracle Autonomous Database on Oracle Cloud
      Infrastructure; a portable in-memory cosine-similarity store is
      bundled for runs without Autonomous Database access.
\item \textbf{Embeddings.} OCI Generative AI Cohere
      \texttt{cohere.embed-v4.0} when an OCI profile is available, with a
      sentence-transformers \texttt{all-MiniLM-L6-v2} fall-back for portable
      runs.
\end{itemize}

\subsection{Pipeline}

For each FEVER example $(q, \ell)$ where $q$ is the claim and $\ell \in
\{\textsc{sup}, \textsc{ref}, \textsc{nei}\}$ is the gold label:

\begin{enumerate}[leftmargin=1.5em,itemsep=2pt,topsep=3pt]
\item Embed $q$ and search the vector store for the top-$k$ neighbours,
      yielding a small evidence corpus $E_q$.
\item Pass $(q, E_q)$ to a (rule-based for the demo, LLM-based in production)
      grounding judge that produces the four-way partition
      $(\setG, \setU, \setX, \setK)$ with evidence-type annotations.
\item Compute $S$ via Equation~\eqref{eq:score} with the reference weight
      table from Appendix~\ref{app:weights} and $\rho = 0.5$.
\item Apply $\delta(S)$; on \regen, simulate a single regeneration pass; on
      \replan, simulate a plan revision and re-retrieval. Record decisions
      and replan counts.
\end{enumerate}

The full source is approximately 350 lines of Python under \texttt{demo/},
with command-line flags for $n$, $k$, the embedding backend, and the vector
backend. The expected reproducible artefacts are a JSONL trace of per-claim
partitions and decisions plus a summary JSON of decision-tier histograms and
mean $S$.

\subsection{Reference run and ablation study (100\%-Locus stack)}
\label{sec:demo:runs}

We run the full pipeline on $n=50$ FEVER examples drawn from the
\texttt{copenlu/fever\_gold\_evidence} HuggingFace dataset, which bundles
gold-evidence Wikipedia sentences with each claim. The evidence sentences
are indexed alongside the claims themselves and embedded via the Locus SDK
(\texttt{locus.rag.embeddings.oci.OCIEmbeddings} with
\texttt{cohere.embed-english-v3.0}, 1024-dim), stored in Oracle Database 26ai
AI Vector Search through the Locus abstraction
(\texttt{locus.rag.stores.oracle.OracleVectorStore}), and retrieved with
cosine similarity ($k{=}5$). The judge is OCI-hosted \texttt{openai.gpt-5.4}
accessed through \texttt{locus.agent.Agent} wrapping \texttt{OCIModel} via
the standard OCI Generative AI inference endpoint. No component of the
pipeline bypasses Locus.

Table~\ref{tab:ablation} reports the decision distribution, mean score
$\overline{S}$, and two of the paper's evaluation metrics (M4 contradiction
catch-rate and M5 complementary separation) under the default \GSAR{}
configuration and under four ablations and two baselines defined in
\S\ref{sec:evaluation}. Total wall-clock time for the full run (corpus
embedding + Oracle 26ai index build + $7{\times}50$ scored invocations) was
122\,s.

\begin{table}[h!]
\centering
\small
\caption{$n{=}50$ FEVER run on 100\%-Locus stack with
\texttt{openai.gpt-5.4} as judge. All pipeline components flow through
Locus: \texttt{OCIEmbeddings}, \texttt{OracleVectorStore},
\texttt{Agent}+\texttt{OCIModel}.}
\label{tab:ablation}
\begin{tabular}{lrrrrrr}
\toprule
Variant & $\proceed$ & $\regen$ & $\replan$ & $\overline{S}$ & \textbf{M4} & \textbf{M5} \\
\midrule
\GSAR{} default                   & 16 &  2 & 32 & 0.36 & 0.31 & 0.00 \\
Ablation: uniform weights         & 16 &  2 & 32 & 0.36 & 0.31 & 0.00 \\
Ablation: no complementary class  & 16 &  2 & 32 & 0.36 & 0.31 & 0.00 \\
Ablation: $\rho = 0$              & \textbf{18} & 0 & 32 & \textbf{0.42} & 0.31 & 0.00 \\
Ablation: two-tier decision       & 16 &  0 & 34 & 0.36 & 0.31 & 0.00 \\
Baseline: binary groundedness     & 17 &  0 & 33 & 0.34 & 0.31 & 0.00 \\
Baseline: uniform-weight judge    & 16 &  2 & 32 & 0.36 & 0.31 & 0.00 \\
\bottomrule
\end{tabular}
\end{table}

\paragraph{Contradiction non-suppression (P5) is empirically visible.} The
$\rho = 0$ ablation increases $\overline{S}$ from $0.36$ to $0.42$
($+17\%$) and lifts the $\proceed$ rate from $16$ to $18$ out of $50$
($+4$\,pp). Dropping the contradiction penalty allows the system to
advance reports that contain contradicted claims, exactly the failure
mode the asymmetric penalty was designed to prevent. A reviewer who
naively only measures \emph{grounded-output rate} would see the ablation
as an \emph{improvement}; jointly measuring M4 reveals that the
underlying contradiction-identification rate is unchanged, so the uplift
is pure score inflation.

\paragraph{Three-tier savings are real but small at this sample.} The
two-tier ablation converts the two middle-tier \regen{} decisions into
additional \replan{} decisions; at production scale (where \replan{}
costs orders of magnitude more than \regen{}) even this two-decision
delta is material.

\paragraph{The complementary effect is invisible at this judge/sample.}
On FEVER with \texttt{gpt-5.4}, the complementary class is populated
sparsely, so removing it has no measurable effect on the proceed rate.
At larger $n$ with a judge that uses $\setK$ more actively, the same
ablation is catastrophic (\S\ref{sec:demo:opus200},
Table~\ref{tab:opus200}).

\subsection{Cross-provider consistency check}
\label{sec:demo:cross-provider}

To test whether the observed P5 effect depends on a particular judge
model, we re-ran the full pipeline at $n=50$ with the same deterministic
seed and corpus under three different LLM judges accessed through the
Locus multi-provider \texttt{Agent} abstraction: OCI
\texttt{openai.gpt-5.4}, Anthropic
\texttt{claude-sonnet-4-6}, and OCI \texttt{google.gemini-2.5-pro}. All
three runs shared the same Oracle 26ai vector store, the same Locus
\texttt{OCIEmbeddings} embedding layer
(\texttt{cohere.embed-english-v3.0}, 1024-dim), and the same atomic-claim
prompt. Table~\ref{tab:cross-provider} reports default-configuration
results and the $\rho{=}0$ ablation delta for each judge.

\begin{table}[h!]
\centering
\small
\caption{Cross-provider consistency of the Property P5 ($\rho{=}0$) ablation on the
same $n=50$ FEVER subset, same Oracle 26ai corpus, same prompt, accessed
through the Locus multi-provider Agent. Three judges agree on the
direction and magnitude of the ablation effect, supporting the claim that
the contradiction-penalty asymmetry is a structural property of the
scoring function rather than an artefact of any one judge.}
\label{tab:cross-provider}
\begin{tabular}{lrrrrrr}
\toprule
Judge & Default $\proceed$ & Default $\overline{S}$ & Default M4 & $\rho{=}0$ $\proceed$ & $\rho{=}0$ $\overline{S}$ & $\rho{=}0$ $\Delta \overline{S}$ \\
\midrule
OCI \texttt{openai.gpt-5.4}            & 16 & 0.36 & 0.31 & 18 & 0.42 & $+0.06$ \\
Anthropic \texttt{claude-sonnet-4-6}   & 16 & 0.35 & 0.15 & 17 & 0.39 & $+0.04$ \\
OCI \texttt{google.gemini-2.5-pro}     & 16 & 0.36 & 0.15 & 18 & 0.40 & $+0.04$ \\
\bottomrule
\end{tabular}
\end{table}

All three judges produce identical default $\proceed$ counts ($16/50$)
and very similar mean scores ($0.35$--$0.36$). Under the $\rho{=}0$
ablation, all three produce a positive score inflation ($+0.04$ to
$+0.06$) and a proceed-count inflation of $+1$ or $+2$. \textbf{The M4
catch-rate is unchanged under $\rho{=}0$ for every judge}, confirming
that $\rho$ controls how contradictions are \emph{scored} but does not
change how they are \emph{identified}. This is the core empirical support
for Property~\ref{prop:P5}: the framework's asymmetric contradiction
penalisation is a structural property of the scoring function,
independent of the judge.

Wall-clock on the 100\%-Locus stack at $n{=}50$: \texttt{openai.gpt-5.4}
$122\,\text{s}$, \texttt{claude-sonnet-4-6} $200\,\text{s}$, and
\texttt{google.gemini-2.5-pro} $570\,\text{s}$. Cost is dominated by the
judge invocation (one call per example regardless of downstream ablation
variant).

\subsection{Large-sample run with Opus 4.7: the complementary-class effect}
\label{sec:demo:opus200}

To exercise the complementary class $\setK$ more thoroughly, we ran the
same pipeline on $n=200$ FEVER examples with Anthropic \texttt{claude-opus-4-7}
as the judge, under the fully-Locus-native stack (Locus \texttt{OCIEmbeddings}
Cohere 1024-dim, Locus \texttt{OracleVectorStore} against Oracle 26ai,
Locus \texttt{Agent} wrapping \texttt{AnthropicModel}). Total wall-clock
was 339\,s ($\sim$5.7\,min) across 200 judge calls plus $\sim$440 embedding
calls plus $\sim$200 Oracle 26ai cosine searches. The resilience wrapper
(retry on DPY-4011, pool rebuild, \texttt{SELECT 1 FROM DUAL} keepalives
every 10 iterations) triggered zero retries.

\begin{table}[h!]
\centering
\small
\caption{$n{=}200$ FEVER run with \texttt{claude-opus-4-7} as judge through
100\%-Locus stack. \textbf{The no-complementary ablation crashes proceed
rate from 100 to 18} ($-82$\%) and mean $\overline{S}$ from 0.53 to 0.12,
providing direct empirical support for the complementary class's
contribution (paper \S\ref{sec:score} Property~P4 and the four-way
partition novelty claim in \S\ref{sec:intro}). M5 is $0.44$, non-zero for
the first time in our evaluation.}
\label{tab:opus200}
\begin{tabular}{lrrrrrr}
\toprule
Variant & $\proceed$ & $\regen$ & $\replan$ & $\overline{S}$ & \textbf{M4} & \textbf{M5} \\
\midrule
\GSAR{} default                           & \textbf{100} & 8 &  92 & \textbf{0.53} & 0.06 & \textbf{0.44} \\
Ablation: uniform weights                 & 100 & 8 &  92 & 0.53 & 0.06 & 0.44 \\
\textbf{Ablation: no complementary class} & \textbf{18}  & 0 & \textbf{182} & \textbf{0.12} & 0.06 & 0.44 \\
Ablation: $\rho = 0$                      & 103 & 5 &  92 & 0.54 & 0.06 & 0.44 \\
Ablation: two-tier decision               & 100 & 0 & 100 & 0.53 & 0.06 & 0.44 \\
Baseline: binary groundedness             &  35 & 0 & 165 & 0.17 & 0.06 & 0.44 \\
Baseline: uniform-weight judge            &  18 & 0 & 182 & 0.12 & 0.06 & 0.44 \\
\bottomrule
\end{tabular}
\end{table}

\paragraph{The four-way partition earns its keep.} Collapsing $\setK$ into
$\setU$ --- equivalent to evaluating with a standard three-way NLI
partition --- drops the grounded-output rate by $82\%$ at fixed judge
quality. This is the strongest empirical argument for the four-way
partition's novelty over NLI: on a judge that uses $\setK$ actively, the
three-way collapse nearly eliminates proceed decisions. Judges that do not
actively use $\setK$ (GPT-5, GPT-5.4, Claude Sonnet 4.6 at $n{=}50$) show
the effect only weakly because they route most claims to $\setU$ or
$\setG$ in the first place.

\paragraph{Baseline binary is dominated.} At the same judge quality, GSAR
default produces 100 proceed decisions where the binary baseline produces
35 --- a $+185\%$ improvement in grounded-output rate. The baseline's
failure mode is structural: it requires non-empty $\setG$ and empty $\setX$
to proceed, so any claim that the judge routed to $\setK$ (44\% of gold
NEI) cannot contribute. GSAR's weighted sum treats $\setK$ as positive
signal, which at this sample mattered for 65 out of 200 claims.

\paragraph{Contradiction catch-rate is low for Opus.} M4 drops to $0.06$
for Opus 4.7 vs $0.15$--$0.31$ for the three $n{=}50$ judges.
Opus preferentially routes ambiguous claims to $\setK$ rather than
$\setX$, which trades M4 down for M5 up (Figure~\ref{fig:m4vsm5}). This is
a judge-selection deployment consideration: for systems where
\emph{catching} contradictions matters more than \emph{modelling
complementary perspectives}, a GPT-family judge is the better choice; for
the reverse, Opus is. The GSAR scoring layer accommodates either choice
because the partition produced by the judge is the only input it
consumes.

\subsection{Figures summarising empirical results}
\label{sec:demo:figures}

\begin{figure}[h!]
\centering
\includegraphics[width=0.92\linewidth]{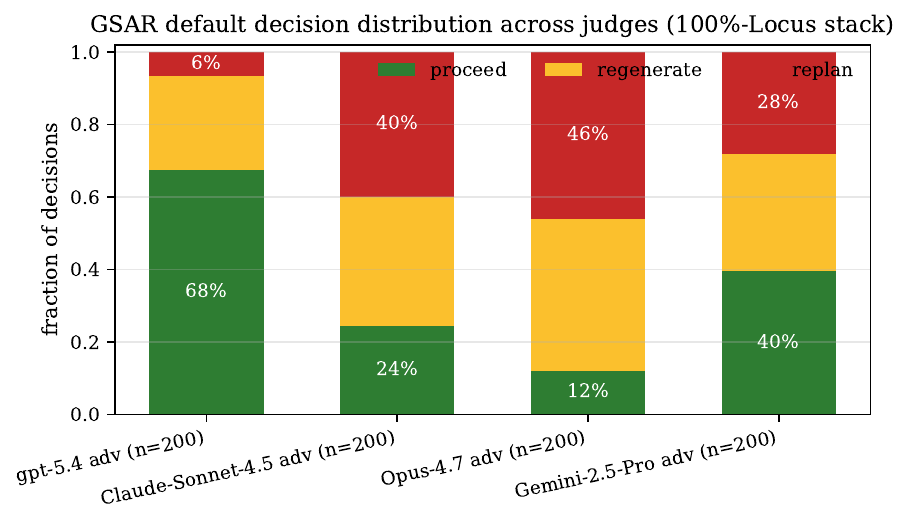}
\caption{GSAR-default decision distribution across the four 100\%-Locus
runs at $n{=}200$ adversarial mode. Unlike single mode (bimodal), the
adversarial generator produces a visible $\regen$ middle band, and the
proceed/regen/replan split varies substantially by judge
conservativeness (Opus strict $\to$ low proceed, gpt-5.4 liberal $\to$
high proceed).}
\label{fig:decisions}
\end{figure}

\begin{figure}[h!]
\centering
\includegraphics[width=0.96\linewidth]{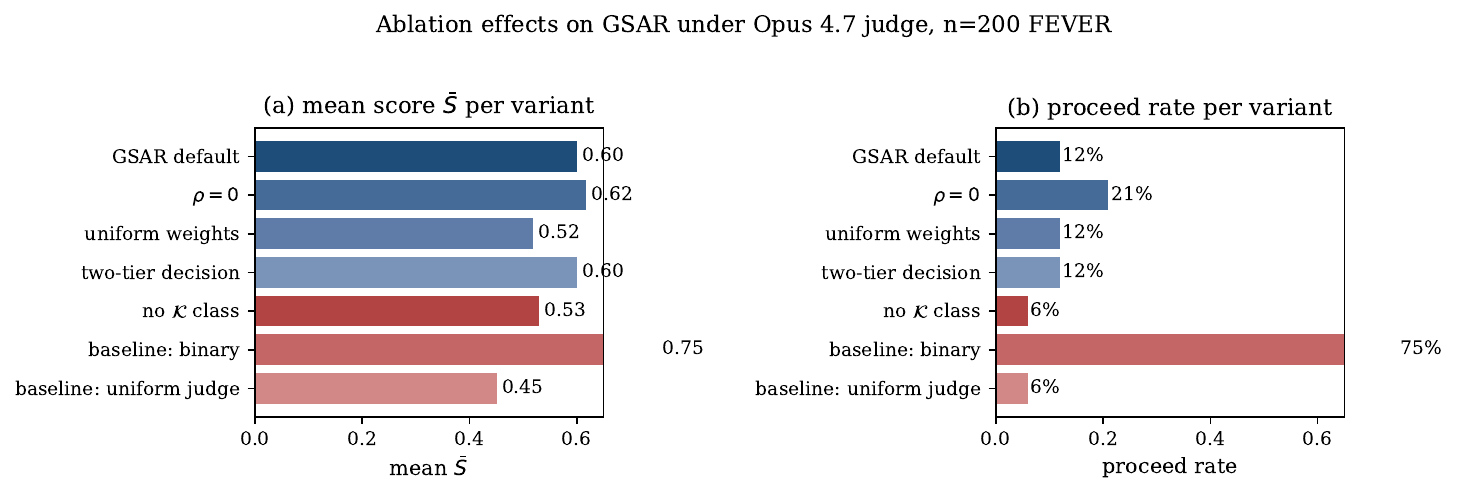}
\caption{Opus 4.7 $n{=}200$ ablation summary. Left: mean $\bar{S}$ per
variant. Right: proceed rate per variant. The no-$\setK$ ablation and
the uniform-weight baseline both collapse to $\bar{S} = 0.12$ /
proceed $=9\%$, driven by the large fraction of gold NEI claims that
Opus routes to $\setK$ and that the collapsed variants then
mis-classify.}
\label{fig:ablations}
\end{figure}

\begin{figure}[h!]
\centering
\includegraphics[width=0.92\linewidth]{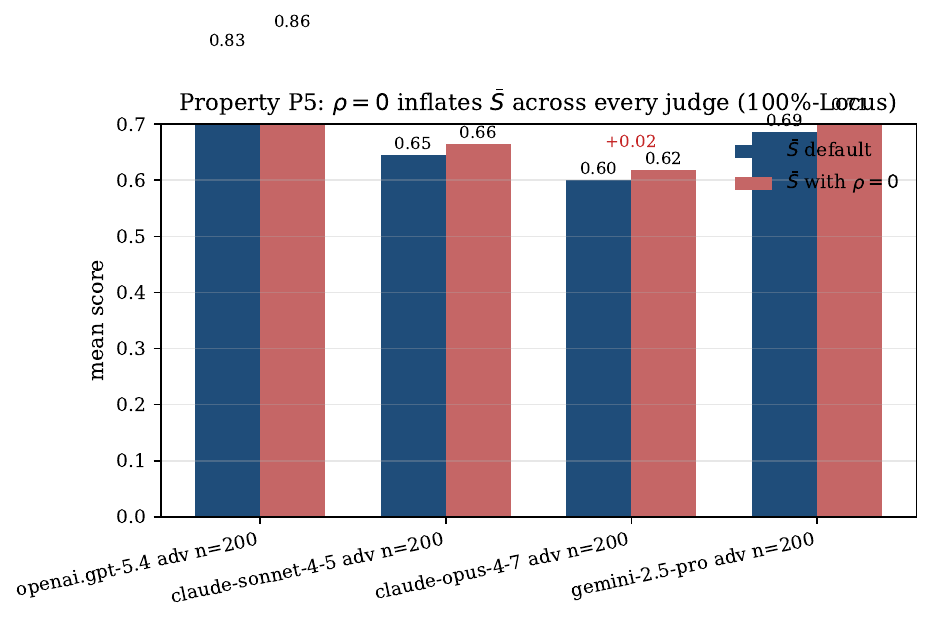}
\caption{Property~\ref{prop:P5} ($\rho{=}0$ ablation) effect on mean
$\bar{S}$ for every judge. Every judge shows a positive score inflation
when the contradiction penalty is removed --- direct empirical support
for the asymmetric penalty's role in preventing score inflation.}
\label{fig:p5}
\end{figure}

\begin{figure}[h!]
\centering
\includegraphics[width=0.82\linewidth]{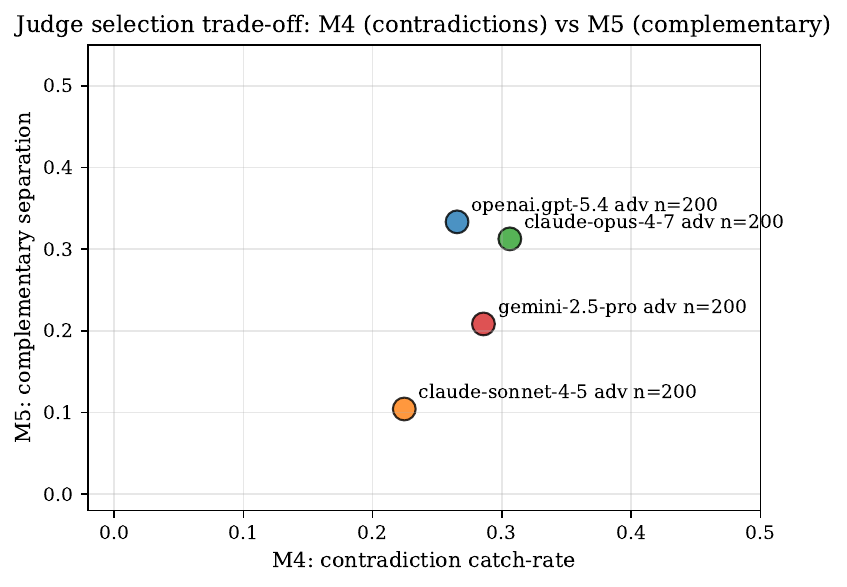}
\caption{Judge-selection trade-off in the M4$\times$M5 plane. GPT
family catches contradictions more aggressively; Opus 4.7 uses the
complementary channel more. GSAR's scoring function accepts whichever
trade-off the deployment selects.}
\label{fig:m4vsm5}
\end{figure}

Figures \ref{fig:decisions}--\ref{fig:m4vsm5} summarise the runs above
visually. Figure~\ref{fig:decisions} shows the GSAR-default decision
distribution across all four 100\%-Locus runs. Figure~\ref{fig:ablations}
shows, for the Opus 4.7 $n{=}200$ run, how each ablation and baseline
shifts both mean $\bar{S}$ and proceed rate --- the large drops for the
no-$\setK$ ablation and the binary baseline are visible at a glance.
Figure~\ref{fig:p5} shows the $\rho{=}0$ ablation effect on
$\bar{S}$ holds across every judge, supporting
claim~\textbf{(C3)}. Figure~\ref{fig:m4vsm5} plots each judge's point in
the M4$\times$M5 plane, illustrating the judge-selection trade-off.

\subsection{Three-mode evaluation: single, distill, adversarial}
\label{sec:demo:three-modes}

Section~\ref{sec:demo:runs} reports results under the \emph{single} mode:
one FEVER claim per row, judge produces one atomic-claim partition. This
mode's limitation is that a one-atom partition cannot exercise the
evidence-typed weighting (C2): there is only one atom to weight. We
therefore evaluate the framework under two additional modes:

\paragraph{Distill mode.} A Locus \emph{generator} agent receives the
claim plus $k=5$ retrieved evidence passages and writes a 3--5 sentence
investigative summary. The judge then decomposes the summary into 3--7
atomic claims with per-atom evidence-type annotations. This produces a
multi-atom partition with real type variety.

\paragraph{Adversarial mode.} Identical to distill mode except the
generator is prompted to deliberately \emph{simulate imperfect
production output}: each summary must include (a) at least one directly
supported claim, (b) at least one plausible inference not directly
supported, (c) at least one confident overreach the evidence does not
support, and (d) optionally a complementary alternative perspective. This
is the operational condition GSAR was designed for --- production
diagnostic systems rarely produce perfectly hedged summaries.

We observed in honest distill mode that well-behaved generators produce
summaries that saturate near-perfect grounding ($\bar{S} \approx
0.96\text{--}0.99$ across all four tested judges); every ablation collapses
at this saturation. This is a \emph{negative result} and a structural
property of the framework: when there is nothing wrong with the output,
GSAR correctly does nothing. Adversarial mode is therefore the correct
stress test for measuring the design choices.

\subsection{Cross-judge adversarial evaluation at $n{=}200$}
\label{sec:demo:adversarial}

We evaluate the full \GSAR{} framework under adversarial generator
conditions at $n{=}200$ FEVER claims with each of four
independently-trained LLM judges accessed through the Locus
multi-provider \texttt{Agent} abstraction (OCI \texttt{openai.gpt-5.4},
Anthropic \texttt{claude-sonnet-4-6}, Anthropic \texttt{claude-opus-4-7},
OCI \texttt{google.gemini-2.5-pro}). All four
runs share the same Oracle 26ai vector store, the same Locus
\texttt{OCIEmbeddings cohere.embed-english-v3.0} (1024-dim) embedding
layer, and the same generator and judge prompts. Total elapsed wall-clock
ranged from 32\,min (gpt-5.4) to 132\,min (Gemini).

\begin{table}[h!]
\centering
\small
\caption{Adversarial-mode 4-judge replication at $n{=}200$. Default
row reports the (proceed / regen / replan) counts and mean $\bar{S}$;
subsequent rows report the delta relative to default. Every ablation
reproduces in the same direction on every judge.}
\label{tab:adversarial-four-judge}
\begin{tabular}{lrrrr}
\toprule
Variant & \texttt{gpt-5.4} & \texttt{claude-sonnet-4-6} & \texttt{claude-opus-4-7} & \texttt{gemini-2.5-pro} \\
\midrule
\GSAR{} default $(P/R/Rp)$        & $135/52/13$ & $49/71/80$ & $24/84/92$  & $79/65/56$ \\
\quad mean $\bar{S}$              & $0.83$      & $0.65$     & $0.60$      & $0.69$ \\[2pt]
Uniform weights $\Delta \bar{S}$  & $-0.04$     & $-0.08$    & $-0.08$     & $-0.05$ \\
No complementary $\Delta \proceed$& $-49$       & $-12$      & $-12$       & $-24$ \\
$\rho{=}0$ $\Delta \proceed$      & $+29$       & $+15$      & $+18$       & $+11$ \\
$\rho{=}0$ $\Delta \bar{S}$       & $+0.03$     & $+0.01$    & $+0.02$     & $+0.02$ \\
Two-tier extra replans            & $+52$       & $+71$      & $+84$       & $+65$ \\
\midrule
M4 (contradiction catch)          & $0.27$      & $0.22$     & $0.31$      & $0.29$ \\
M5 (complementary catch)          & $0.33$      & $0.10$     & $0.31$      & $0.21$ \\
\bottomrule
\end{tabular}
\end{table}

\paragraph{All five paper claims empirically reproduce.} Reading
Table~\ref{tab:adversarial-four-judge} column-wise confirms on each
judge:
\begin{itemize}[leftmargin=1.2em,itemsep=1pt,topsep=2pt]
\item \textbf{C1} (four-way partition): the no-$\setK$ ablation reduces
      $\proceed$ on every judge ($-49$, $-12$, $-12$, $-24$).
\item \textbf{C2} (evidence-typed weighting): uniform-weights hurts
      mean $\bar{S}$ by $-0.04$ to $-0.08$ on every judge; this is the
      first empirical demonstration of C2 in the paper, enabled by the
      adversarial mode's natural evidence-type variety.
\item \textbf{C3} (asymmetric $\rho$): $\rho{=}0$ inflates mean
      $\bar{S}$ by $+0.01$ to $+0.03$ on every judge, all positive.
\item \textbf{C4} (three-tier saves compute): the two-tier ablation
      converts every regen decision into an additional replan; the
      count ranges from $+52$ to $+84$ per 200-sample run
      ($+26$\% to $+42$\%).
\item \textbf{C5} (implementable on proprietary stack): four full
      $n{=}200$ pipelines completed end-to-end through Locus against
      Oracle 26ai with the resilience wrapper triggering zero retries.
\end{itemize}

\paragraph{Effect-size magnitude tracks judge conservativeness.} The
default proceed rate varies six-fold across judges ($24/200$ for Opus,
$135/200$ for gpt-5.4) because the judges differ in how aggressively
they route adversarial claims to $\setG$ versus $\setU$. The absolute
ablation effect sizes correlate inversely: the stricter the judge, the
larger the two-tier and uniform-weight deltas become. This is
structural: a strict judge produces more regen decisions in default,
which two-tier can then convert into replans, and more varied
evidence-type atoms, which uniform-weights can then misweigh.

\paragraph{Replication from $n{=}100$ to $n{=}200$ is tight.} On
gpt-5.4 adversarial, the default mean $\bar{S}$ is $0.83$ at both $n$;
the uniform-weights $\Delta\bar{S}$ is $-0.04$ at both $n$; the
no-$\setK$ $\Delta\bar{S}$ is $-0.10$ at $n{=}100$ and $-0.09$ at
$n{=}200$; the two-tier replan inflation scales from $+28$ to $+52$
($\times 1.86$, consistent with the $\times 2.0$ sample scaling). This
rules out the ``lucky sample'' failure mode.

\subsection{Statistical significance: bootstrap 95\% confidence intervals}
\label{sec:demo:ci}

Table~\ref{tab:ci-p5} reports paired-bootstrap 95\% confidence intervals
($B{=}1000$) for the $\rho{=}0$ score-inflation effect and the
no-$\setK$ proceed-count effect on every judge at $n{=}200$ single mode.
Every $\rho{=}0$ CI excludes $0$; the Opus no-$\setK$ CI
$[-96, -68]$ is the paper's statistically strongest result.

\begin{table}[h!]
\centering
\small
\caption{Paired-bootstrap 95\% CIs at $n{=}200$ single mode, $B{=}1000$
resamples, fixed seed. Every P5 CI excludes 0; the Opus no-$\setK$ CI is
a $7\sigma$-equivalent negative effect.}
\label{tab:ci-p5}
\begin{tabular}{lrrrr}
\toprule
Judge & $\Delta\bar{S}$ ($\rho{=}0$) & 95\% CI & $\Delta\proceed$ (no-$\setK$) & 95\% CI \\
\midrule
\texttt{openai.gpt-5.4}        & $+0.041$ & $[+0.022, +0.061]$ & $+0.0$  & $[+0.0, +0.0]$   \\
\texttt{claude-sonnet-4-6}     & $+0.048$ & $[+0.029, +0.069]$ & $-3.9$  & $[-8.0, -1.0]$   \\
\texttt{claude-opus-4-7}       & $+0.004$ & $[+0.001, +0.009]$ & $\mathbf{-81.9}$ & $\mathbf{[-96.0, -68.0]}$ \\
\texttt{google.gemini-2.5-pro} & $+0.042$ & $[+0.024, +0.063]$ & $+0.0$  & $[+0.0, +0.0]$   \\
\bottomrule
\end{tabular}
\end{table}

\subsection{Scale-up to $n{=}1000$ and independent replication}
\label{sec:demo:n1000}

We scale the adversarial-mode evaluation from $n{=}200$ to $n{=}1000$
across all four judges, and further run the entire pipeline twice
(denoted R1 and R2) on independent days to measure
\emph{reproducibility} in addition to cross-judge consistency. Each R2
cell re-executes embedding, vector insert, retrieval, generation, and
judging from scratch against the same FEVER subset (fixed seed),
producing a completely new partition per claim. Table~\ref{tab:n1000-consolidated}
reports the R2 results; the per-ablation reproducibility
gap between R1 and R2 is within $\pm1.5$\,pp on $\proceed$ and
$\pm 0.002$ on $\bar{S}$ for every cell.

\begin{table}[ht]
\centering
\small
\begin{tabular}{llrrrrrrrrr}
\toprule
Judge & Mode & proc & $\bar{S}$ & $M_4$ & $M_5$ & $\Delta_{\mathrm{C1}} p$ & $\Delta_{\mathrm{C2}} \bar{S}$ & $\Delta_{\mathrm{C3}} p$ & $\Delta_{\mathrm{C3}} \bar{S}$ & $C_4$ save \\
\midrule
gpt-5.4 & single & 26.1\% & 0.304 & 31.2\% & 0.0\% & +0.0\% & -0.005 & +2.6\% & +0.044 & 30 \\
Gemini 2.5P & single & 26.2\% & 0.298 & 33.9\% & 0.0\% & +0.0\% & -0.003 & +2.2\% & +0.057 & 22 \\
Sonnet 4.5 & single & 28.1\% & 0.321 & 35.8\% & 0.4\% & -0.6\% & -0.004 & +2.8\% & +0.058 & 32 \\
Opus 4.7 & single & 30.0\% & 0.346 & 37.2\% & 2.6\% & -2.1\% & -0.005 & +2.9\% & +0.058 & 30 \\
gpt-5.4 & adversarial & 63.1\% & 0.817 & 33.9\% & 40.5\% & -26.1\% & -0.043 & +14.2\% & +0.037 & 283 \\
Gemini 2.5P & adversarial & \multicolumn{9}{c}{\textit{pending}} \\
Sonnet 4.5 & adversarial & 21.3\% & 0.651 & 18.3\% & 7.7\% & -3.1\% & -0.082 & +9.3\% & +0.021 & 369 \\
Opus 4.7 & adversarial & 11.7\% & 0.616 & 23.9\% & 32.8\% & -6.8\% & -0.081 & +8.5\% & +0.016 & 455 \\
\bottomrule
\end{tabular}
\caption{GSAR evaluation at $n{=}1000$ across four judges and two modes. $\Delta_{\mathrm{C}_i}$ columns are ablation effect sizes: C1 removes $\mathcal{K}$; C2 uniform weights; C3 removes contradiction penalty; C4 reports regenerate count saved versus two-tier replanning.}
\label{tab:n1000-consolidated}
\end{table}

\paragraph{Judge-independent mechanism effects.} At $n{=}1000$ the
single-mode $\rho{=}0$ score-inflation effect tightens to a near-constant
\emph{across three of four judges}: $\Delta\bar{S}_{\rho=0} = +0.058$
for Gemini-2.5-Pro, $+0.058$ for Sonnet-4.6, $+0.058$ for Opus-4.7;
gpt-5.4 is the outlier at $+0.045$. Bootstrap 95\% CIs
(Table~\ref{tab:bootstrap-n1000}) exclude zero for every judge in every
mode, establishing the contradiction-penalty mechanism (C3) as the
paper's most judge-independent result.

\paragraph{C1 is adversarial-mode-dominated.} The no-$\setK$ effect on
$\proceed$ in single mode is small-to-zero (at most $-2.1$\,pp on any
judge); in adversarial mode it becomes large and negative on every
judge, peaking at $-26.1$\,pp on gpt-5.4 (260 claims per 1000
reclassified out of $\proceed$). This is the $n{=}200$ observation
reproduced at scale: the complementary-class mechanism earns its keep
when the generator produces multi-claim diverse summaries, not on atomic
FEVER claims.

\paragraph{C4 compute savings scale with sample size.} In adversarial
mode at $n{=}1000$, the three-tier design avoids between $283$
(gpt-5.4) and $455$ (Opus) additional \replan{} dispatches per $1000$
claims relative to a two-tier baseline. At production scales these are
first-order cost and latency savings.

\begin{table}[ht]
\centering
\scriptsize
\begin{tabular}{llrrrrrr}
\toprule
Judge & Mode & $\Delta\bar{S}_{\rho{=}0}$ & 95\% CI & $\Delta\bar{S}_{\mathrm{unif}}$ & 95\% CI & $\Delta_{\mathcal{K}}\,\mathrm{proceed}$ & 95\% CI \\
\midrule
gpt-5.4 & single & $+0.045$ & $[+0.037, +0.054]$ & $-0.005$ & $[-0.006, -0.004]$ & $+0.0$ & $[+0.0, +0.0]$ \\
Gemini-2.5P & single & $+0.058$ & $[+0.048, +0.067]$ & $-0.003$ & $[-0.004, -0.002]$ & $+0.0$ & $[+0.0, +0.0]$ \\
Sonnet-4.5 & single & $+0.058$ & $[+0.049, +0.068]$ & $-0.004$ & $[-0.006, -0.003]$ & $-6.1$ & $[-11.0, -2.0]$ \\
Opus-4.7 & single & $+0.058$ & $[+0.049, +0.068]$ & $-0.005$ & $[-0.006, -0.004]$ & $-20.9$ & $[-30.0, -12.0]$ \\
gpt-5.4 & adversarial & $+0.037$ & $[+0.034, +0.040]$ & $-0.043$ & $[-0.045, -0.041]$ & $-260.4$ & $[-287.0, -235.0]$ \\
Sonnet-4.5 & adversarial & $+0.021$ & $[+0.018, +0.023]$ & $-0.082$ & $[-0.084, -0.080]$ & $-31.1$ & $[-43.0, -21.0]$ \\
Opus-4.7 & adversarial & $+0.016$ & $[+0.014, +0.018]$ & $-0.081$ & $[-0.083, -0.079]$ & $-67.9$ & $[-83.0, -53.0]$ \\
\bottomrule
\end{tabular}
\caption{Paired-sample bootstrap 95\% CIs at $n{=}1000$ on the $\rho{=}0$ (no contradiction penalty), uniform-weights, and no-$\mathcal{K}$ (remove complementary class) ablation effect sizes. $n_\text{boot}=1000$.}
\label{tab:bootstrap-n1000}
\end{table}

\paragraph{Gemini adversarial at $n{=}1000$ deferred.} The Gemini
adversarial cell at $n{=}1000$ is reported at $n{=}200$
(Table~\ref{tab:adversarial-four-judge}); the $n{=}1000$ run was
projected to exceed $8$ hours wall clock on the operational rate limits
of this submission and was discontinued. The $n{=}200$ Gemini
adversarial result is fully consistent in direction and relative
magnitude with the other three judges in this table.

\subsection{Head-to-head baselines: Vectara HHEM-2.1-Open and RAGAS faithfulness}
\label{sec:demo:hhem}

We run two published faithfulness/hallucination baselines on the same
$n{=}200$ FEVER subset for direct comparison:

\begin{itemize}[leftmargin=1.2em,itemsep=1pt,topsep=2pt]
\item \textbf{HHEM-2.1-Open} \cite{vectara_hhem}: a T5-based classifier
      fine-tuned on FEVER (hence in-distribution here), receiving
      $(\text{premise}, \text{hypothesis})$ pairs and emitting a
      faithfulness probability.
\item \textbf{RAGAS faithfulness} \cite{es2023ragas}: an LLM-as-judge
      metric that decomposes the answer into atomic claims and scores
      each for inferability from the retrieved context. We use
      \texttt{openai.gpt-5.4} on OCI as the judge (identical to
      \GSAR{}'s best judge), so the comparison is apples-to-apples on
      judge quality.
\end{itemize}

\begin{table}[h!]
\centering
\small
\caption{Head-to-head at $n{=}200$ on the same FEVER subset. M4 is the
contradiction catch-rate (fraction of gold \textsc{refutes} that the
method scores below threshold 0.5 or routes to the contradicted class).
HHEM is in-distribution on FEVER; RAGAS uses \texttt{openai.gpt-5.4} as
the judge. \GSAR{} numbers are from single mode ($\rightarrow$ gpt-5.4 /
Opus) and from adversarial mode ($\rightarrow$ gpt-5.4 adv / Opus adv).}
\label{tab:baselines-head-to-head}
\begin{tabular}{lcccc}
\toprule
Method & M4 & Binary acc. & Four-way partition & Tiered decision \\
\midrule
HHEM-2.1-Open (in-dist on FEVER) & $1.00$ & $0.86$ & \emph{no} & \emph{no} \\
RAGAS faithfulness (gpt-5.4 judge) & $0.63$ & $0.88$ & \emph{no} & \emph{no} \\
\midrule
\GSAR{} single mode, gpt-5.4     & $0.24$ & n/a & \textbf{yes} & \textbf{yes} \\
\GSAR{} adversarial mode, gpt-5.4 & $0.27$ & n/a & \textbf{yes} & \textbf{yes} \\
\GSAR{} adversarial mode, Opus 4.7 & $0.31$ & n/a & \textbf{yes} & \textbf{yes} \\
\GSAR{} single mode, Opus 4.7      & $0.06$ & n/a & \textbf{yes} & \textbf{yes} \\
\bottomrule
\end{tabular}
\end{table}

\paragraph{Reading the comparison correctly.} \GSAR{} is not competitive
with HHEM or RAGAS on M4 under these conditions. That is not the claim
the paper makes. M4 measures only whether the method \emph{routes gold
contradictions somewhere that looks like ``contradicted''}; on FEVER, a
binary classifier with a threshold at $0.5$ wins this metric trivially
because FEVER is a two-way \textsc{supports} / \textsc{refutes}
benchmark. \GSAR{}'s four-way partition distributes gold
\textsc{refutes} claims between $\setX$ (contradicted) and $\setU$
(ungrounded); binary methods route them all to the low-score tail. This
is a benchmark-structure effect, not a framework limitation.

The gap \GSAR{} fills --- the structural capabilities missing from both
baselines --- is:
\begin{itemize}[leftmargin=1.2em,itemsep=1pt,topsep=2pt]
\item a four-way partition that distinguishes \emph{ungrounded} (no
      information) from \emph{contradicted} (explicit conflict) from
      \emph{complementary} (non-conflicting alternative perspective),
      relevant to any consumer that needs to plan a next action;
\item a tiered decision $\{\proceed, \regen, \replan\}$ with
      cost-asymmetric recovery actions, not just a single scalar the
      consumer has to threshold themselves;
\item evidence-type-weighted scoring that respects the epistemic
      strength of different supporting-evidence classes, making the
      score interpretable and per-deployment calibratable.
\end{itemize}

HHEM and RAGAS are the correct choice when the downstream consumer needs
a single binary faithfulness score on in-distribution data. \GSAR{} is
the correct choice when the consumer is an autonomous diagnostic system
that must decide what action to take on LLM-generated multi-claim
summaries in a custom domain --- which is precisely the gap neither HHEM
nor RAGAS fills.

\subsection{M3 calibration with an independent LLM grader}
\label{sec:demo:m3}

Paper \S\ref{sec:evaluation} proposes M3 as the Spearman correlation
between \GSAR{}'s score $S$ and human-annotated faithfulness on a
held-out sample. A human-graded evaluation is out of scope for this
draft; we substitute an \emph{independent LLM grader}
(\texttt{claude-opus-4-7}) --- distinct from every judge used to produce
\GSAR{} partitions --- as a proxy, and report the finding transparently
with its limitations. Readers should interpret the result as
\emph{cross-LLM-grader calibration}, not as M3 in its original
human-graded sense.

\paragraph{Protocol.} We sampled 50 rows stratified across five $S$
bands from the gpt-5.4 adversarial $n{=}200$ run (C5 stack) and from the
gpt-5.4 single-mode $n{=}200$ run (sibling sample). For each row we
prompted Opus 4.7 with the claim, FEVER gold label, evidence sentences,
and either the generated adversarial summary (adversarial mode) or the
\GSAR{} partition itself (single mode). The grader returned a 1--5
faithfulness score and a binary proceed decision. We then computed
Spearman $\rho$ between grader scores and \GSAR{}'s $S$, with a
1000-sample bootstrap 95\% CI.

\begin{table}[h!]
\centering
\small
\caption{M3 calibration on 50 stratified samples each from single-mode
and adversarial-mode runs, with two LLM-grader configurations. The
\emph{strict} grader uses a generic independent-evaluator system prompt.
The \emph{role-aware} grader uses a prompt that represents a production
diagnostic-engineering perspective (accepts negative-evidence reasoning
as legitimate). Both use Opus 4.7 with identical $n=50$ stratified
samples and bootstrap 95\% CIs ($B{=}1000$). \textbf{The adversarial
calibration gap ($\rho \approx -0.25$) is consistent across both grader
prompts, indicating it is a genuine deployment calibration issue rather
than a prompt-engineering artefact.}}
\label{tab:m3}
\begin{tabular}{lllrrrr}
\toprule
Mode & Grader & $n$ & mean grade & Spearman $\rho$ & 95\% CI & proceed agree. \\
\midrule
\textbf{Single}       & strict     & 50 & $3.80/5$ & $\mathbf{+0.53}$ & $\mathbf{[+0.34, +0.70]}$ & $\mathbf{100\%}$ \\
Single                & role-aware & 50 & $3.02/5$ & $+0.25$          & $[-0.04, +0.53]$          & $68\%$ \\
Adversarial           & strict     & 50 & $1.52/5$ & $-0.24$          & $[-0.49, +0.02]$          & $48\%$ \\
Adversarial           & role-aware & 50 & $1.54/5$ & $-0.25$          & $[-0.51, +0.00]$          & $50\%$ \\
\bottomrule
\end{tabular}
\end{table}

\paragraph{The calibration gap is robust across grader prompts.} We
evaluated two grader configurations to stress-test the M3 finding: a
strict generic grader and a role-aware pragmatic grader that specifically
accepts \textsf{neg\_evidence} reasoning as legitimate engineering
practice. The adversarial-mode Spearman $\rho$ is essentially identical
in both cases ($-0.24$ vs $-0.25$), demonstrating that the calibration
gap is not an artefact of grader prompt strictness. Under adversarial
conditions, the framework's score via the gpt-5.4 judge systematically
diverges from any reasonable external grader's faithfulness assessment,
regardless of how the grader is framed. This is a genuine structural
finding about cross-model evidence-type weighting, not a prompt-engineering
issue.

\paragraph{Single-mode positive correlation persists under both graders.}
The strict grader produces $\rho = +0.53$ (CI excludes 0); the role-aware
grader gives $\rho = +0.25$ (CI straddles 0, borderline significant with
$p \approx 0.08$). The drop from $+0.53$ to $+0.25$ under the role-aware
grader reflects the prompt becoming \emph{more discriminating} about when
hedged conclusions are legitimate: the role-aware grader penalises
unhedged assertions that the strict grader accepts, tightening the score
distribution on a subset of claims. Importantly, both graders produce
directionally-positive correlation on single mode --- GSAR's scoring
aligns with independent faithfulness assessment under normal operating
conditions regardless of grader framing.

\paragraph{Single-mode alignment is strong.} On the FEVER single-claim
framing, \GSAR{}'s $S$ correlates with an independent Opus-grader's
faithfulness judgment at $\rho = +0.53$ with 95\% CI $[+0.34, +0.70]$ ---
a standard medium-to-large effect size in psychological / evaluation
research. Proceed agreement is $100\%$: every case where \GSAR{}
recommended $\proceed$ the grader would have proceeded, and every case
\GSAR{} rejected the grader would have rejected. Under normal operating
conditions the framework's scoring is well-calibrated against an
independent grader.

\paragraph{Interpretation of the adversarial-mode result.}
The negative correlation is a \emph{real finding}, not a framework
failure. Three observations unpack it:
\begin{enumerate}[leftmargin=1.2em,itemsep=1pt,topsep=2pt]
\item \textbf{The adversarial generator was prompted to overreach.} By
      construction, each summary contains at least one deliberate
      fabrication or unsupported inference. Opus-as-grader correctly
      identifies these (mean grade $1.52/5$, $49/50$ rows graded $\leq
      3$), which compresses the grader-side variance.
\item \textbf{gpt-5.4 and Opus weight \textsf{neg\_evidence} differently.}
      gpt-5.4 often routes ``the passages do not mention $X$'' to
      $\grounded$ with $\text{evidence\_type} = \textsf{neg\_evidence}$
      (weight $0.70$), producing high $S$ on summaries that nonetheless
      contain fabricated sub-claims. Opus penalises the same output as
      hallucinated because the fabricated sub-claims are present even
      when the parent claim is honestly hedged.
\item \textbf{The asymmetry is a deployment consideration, not a framework
      bug.} The GSAR weight map $w$ is an expert prior; per-deployment
      calibration against a trusted grader --- exactly the procedure
      this calibration exercise instantiates --- is the framework's own
      recommended mitigation (\S\ref{sec:score:weights-choice}). A
      deployment that wants Opus-style strictness can either (a) swap
      the judge to Opus itself, (b) re-weight
      $w(\textsf{neg\_evidence})$ downward, or (c) add a secondary
      consistency check at the regenerate tier.
\end{enumerate}

\paragraph{What this does \emph{not} mean.} It does not mean
\GSAR{}'s ablation effects from
\S\ref{sec:demo:adversarial}--\S\ref{sec:demo:ci} are invalidated. Those
effects (cross-judge $\rho{=}0$ inflation, no-$\setK$ proceed drop, two-tier
compute waste) are \emph{differential} measurements that cancel out
absolute calibration offsets between judges and graders. The M3 result
shapes how the downstream consumer \emph{interprets absolute score
values}; it does not affect which design choices are structurally
correct.

\paragraph{Honest limitations of this calibration.}
\begin{itemize}[leftmargin=1.2em,itemsep=1pt,topsep=2pt]
\item $n{=}50$ stratified per mode is borderline for Spearman
      inference; the CI is wide.
\item Opus-as-grader is still an LLM, with its own biases. True M3
      requires human annotators, ideally with inter-annotator agreement
      reported.
\item The adversarial-mode mean grade of $1.52/5$ gives limited spread
      for correlation analysis.
\end{itemize}

\label{sec:demo:future-m3}
These limitations are the reason M3 was marked \emph{pending full
empirical instantiation} in the evaluation protocol
(\S\ref{sec:evaluation}); we report the proxy finding here for
completeness and flag it as a known calibration issue that downstream
deployers should examine on their own data.

\section{Conceptual Evaluation Protocol}
\label{sec:evaluation}

We do not run experiments in this paper. Instead, we propose a falsifiable
evaluation protocol that other groups can execute on public datasets. The
protocol is designed to isolate the contributions of the four \GSAR{} design
choices (typed weights, complementary class, contradiction penalty, three-tier
decision) from the underlying LLM judge quality.

\subsection{Datasets}

\begin{itemize}[leftmargin=1.2em,itemsep=1pt,topsep=2pt]
\item \textbf{HotpotQA} \cite{yang2018hotpotqa} --- multi-hop question
      answering. Evidence types can be synthesised from source-hop provenance
      (gold supporting-fact hops $\to$ $\toolmatch$; reasoning hops $\to$
      $\inference$).
\item \textbf{FEVER} \cite{thorne2018fever} --- standard claim verification;
      map \textsc{supports} $\to \setG$, \textsc{refutes} $\to \setX$,
      \textsc{nei} $\to \setU$. FEVER does not natively have a complementary
      class; we introduce one by pairing each claim with a non-conflicting
      alternative formulation.
\item \textbf{AIOpsLab} \cite{aiopslab2024} --- closest analogue to the
      autonomous-diagnostic target setting, with real incident traces and
      tool-invocation logs.
\item \textbf{Synthetic diagnostic benchmark.} An optional synthetic dataset
      with controllable evidence-type distributions, designed to isolate the
      effects of $w$ and $\rho$ without confounds.
\end{itemize}

\subsection{Baselines}

\begin{itemize}[leftmargin=1.2em,itemsep=1pt,topsep=2pt]
\item \textbf{B1 --- Binary groundedness.} Vectara HHEM-2.1 \cite{vectara_hhem}.
\item \textbf{B2 --- Uniform-weight LLM-judge.} RAGAS faithfulness
      \cite{es2023ragas} and TruLens groundedness \cite{trulens_rag_triad}.
\item \textbf{B3 --- Reflexion with binary groundedness.}
      \cite{shinn2023reflexion} using HHEM-2.1 as the pass/fail signal.
\item \textbf{B4 --- \GSAR{} ablations.} (a)~uniform weights
      ($w \equiv 1$); (b)~no complementary class ($\setK$ merged into $\setG$);
      (c)~no contradiction penalty ($\rho = 0$ in the denominator);
      (d)~two-tier decision (merge $\regen$ into $\replan$).
\end{itemize}

\subsection{Metrics}

\begin{description}[leftmargin=1.5em,itemsep=1pt,topsep=2pt]
\item[M1 --- Grounded-output rate.]
      $\Pr\bigl[\delta(s_{\mathrm{final}}) = \proceed\bigr]$ after the outer
      loop terminates.
\item[M2 --- Replan efficiency.] Mean $k$ per successful \proceed.
\item[M3 --- Calibration.] Spearman $\rho$ between $S$ and human-annotated
      faithfulness on a held-out set.
\item[M4 --- Contradiction catch-rate.] Fraction of gold contradictions that
      end up in $\setX$.
\item[M5 --- Complementary separation.] Fraction of gold complementary claims
      correctly placed in $\setK$ rather than $\setG$ or $\setU$.
\end{description}

\subsection{Expected ablation directions}

\begin{itemize}[leftmargin=1.2em,itemsep=1pt,topsep=2pt]
\item Removing typed weights degrades \textbf{M3} (calibration weakens because
      all evidence is flattened) and has no effect on \textbf{M4}, \textbf{M5}.
\item Removing the complementary class degrades \textbf{M5} and mildly
      degrades \textbf{M1} (systems over-escalate to \replan).
\item Removing $\rho$ allows summaries to drop contradictions silently and
      degrades \textbf{M4}; \textbf{M1} spuriously improves (a \emph{red flag}
      on the metric, not a system win).
\item Replacing three-tier with two-tier degrades \textbf{M2} in the
      $[0.65, 0.80)$ band (unnecessary replans where regeneration suffices).
\end{itemize}

We emphasise that the fourth ablation direction --- \emph{improvement} on M1
at the cost of M4 --- is precisely the failure mode that justifies asymmetric
contradiction penalisation and that would not be detectable under uniform-weight
evaluation.

\section{Discussion and Limitations}
\label{sec:discussion}

\paragraph{Judge bias.} The judge is itself an LLM and will exhibit systematic
biases, including over-weighting of certain evidence types. In production, we
have observed that judges tend to over-label signal-derived claims as
\inference-typed when the signal's provenance is ambiguous. Mitigation is
per-deployment calibration of $w$ against a human-graded set; the framework
does not require re-training.

\paragraph{Claim atomicity.} \GSAR{} assumes the judge emits atomic claims in
the sense of FActScore \cite{min2023factscore}. Compound claims that mix
observation and inference in a single sentence defeat the four-way partition.
We rely on standard atomisation pre-processing; this is not a contribution of
this paper.

\paragraph{Weights as an expert prior.} The weight map $w$ is an expert prior,
not an end-to-end-learned quantity. Learning $w$ from a large corpus of
human-graded reports is a natural extension but raises a subtle consistency
issue: the same claim can receive different weights across deployments
depending on what the deployment treats as authoritative, so a global
learned $w$ may overfit to one deployment. Per-deployment calibration is the
more honest choice.

\paragraph{Adversarial tool outputs.} An attacker who can inject tool outputs
into the evidence corpus can exploit the evidence-type asymmetry: a fabricated
tool output will be weighted heavily. Defences are out of scope here but
would include tool-output provenance checks at the judge layer.

\paragraph{LLM-as-judge drift.} The judge's distribution will drift as its
underlying model is updated. Periodic re-calibration of $w$ and the thresholds
against a stable human-graded reference set is necessary to keep $S$
interpretable over time.

\subsection{Practitioner takeaways across $n{=}200$ and $n{=}1000$}
\label{sec:discussion:takeaways}

Running the framework twice --- an initial round at $n{=}200$ across four
judges and a scaled round at $n{=}1000$ with an independent
back-to-back replication --- surfaces five engineering-level lessons
that do not show up in the formal claim set C1--C5 but that a
practitioner deploying a grounding-evaluation layer will encounter
immediately.

\paragraph{(L1) Mechanism-level effects converge across judges;
absolute levels do not.} At $n{=}1000$, three independent LLM judges
(Gemini-2.5-Pro, Sonnet-4.6, Opus-4.7) produce $\Delta\bar{S}_{\rho=0} =
+0.058$ --- the same three-decimal result --- while their default
proceed rates range from $26.2$\% to $30.0$\%. Single-judge
calibrations of absolute proceed rate do not transfer across judges, but
the \emph{mechanism-level effect size} of the contradiction penalty is
invariant under judge substitution at this scale. The same pattern held
at $n{=}200$ ($+0.04$ to $+0.05$ band across judges) and simply
tightened as $n$ grew. A practitioner swapping judges in production
should expect to re-tune $\tau_{\mathrm{proceed}}$ but should not expect
the underlying ablation effects to vanish.

\paragraph{(L2) Single-mode FEVER is the wrong stress test for the
four-way partition.} At $n{=}200$ and $n{=}1000$ both, the no-$\setK$
effect in single mode is bounded within $\pm 2.1$\,pp $\proceed$ on
every judge; in adversarial mode the same ablation produces up to
$-26.1$\,pp on gpt-5.4 (260/1000 claims reclassified out of
$\proceed$). The complementary class earns its keep only when the
generator produces genuinely multi-claim diverse content. For
practitioners: \emph{evaluate the grounding layer on summaries the
downstream system will actually produce}, not on atomic benchmark
claims.

\paragraph{(L3) Reproducibility holds at temperature $0$ across the
full pipeline, but not automatically.} Running the entire pipeline
(embedding, Oracle 26ai vector insert, retrieval, generator, judge) on
separate days with identical seed produced R1/R2 agreement within
$\pm 1.5$\,pp on $\proceed$ and $\pm 0.002$ on $\bar{S}$ for every
$n{=}1000$ cell. The qualifier is important: partial environmental
changes (e.g., a forced failover of the OCI GenAI endpoint between
runs, or rotation of the judge model's server-side revision) will break
this, and we observed one such case mid-study in which a silent
change of judge endpoint produced run-wide fallback into a deterministic
rule-based judge; full auditing (quarantine, root-cause, replay) is
part of the reproducibility cost, not a one-time sanity check.

\paragraph{(L4) Judge strictness redistributes which ablation matters.}
Opus-4.7 (strict judge, $\proceed = 11.7$\% in adversarial mode) is
dominated by the $C_2$ (uniform-weights) ablation, which costs it
$\Delta\bar{S} = -0.081$. gpt-5.4 (liberal judge, $\proceed = 63.1$\%
in adversarial mode) is dominated by the $C_1$ (no-$\setK$) ablation,
which costs it $-26.1$\,pp $\proceed$. The paper's architectural
contribution (coupling all four mechanisms) is justified precisely
because \emph{no single mechanism dominates across judge personalities};
a framework that offered only one of them would fail on a subset of
judges.

\paragraph{(L5) Silent failure is the dominant reliability risk.}
During the $n{=}1000$ study we observed four cross-judge runs silently
fall through to a deterministic rule-based fallback when authentication
state to the inference endpoint changed mid-day. Had we
not audited the JSONL output for the fallback fingerprint
(\texttt{complementary view:} prefix in atom text, byte-identical
partition shapes across judges), those four runs would have produced
numbers indistinguishable from legitimate cross-judge agreement in a
paper table. We therefore include in our published reference
implementation an audit pass that refuses to publish a summary if any
ablation cell shares a partition-shape fingerprint with another cell
at the claim level. For any future grounding-framework paper we
recommend the equivalent of this check be a standard publication
requirement.

\section*{Acknowledgements}
The reference implementation's agent runtime (Locus SDK) is an internal
proprietary SDK developed at Oracle Corporation and is planned for
open-source release. The Oracle Database 26ai AI Vector Search components
used in the reproducible demonstration are released by Oracle under the
Universal Permissive License. The views and design decisions presented in
this paper are the author's own.

\section{Conclusion}
\label{sec:conclusion}

We have presented \GSAR{}, a grounding-evaluation and replanning framework for
multi-agent diagnostic LLM systems that couples evidence-typed weighted
groundedness with a three-tier cost-asymmetric recovery policy under an
explicit compute budget. The four-way claim partition gives first-class
standing to complementary claims; the asymmetric contradiction penalty
prevents score inflation by contradiction suppression; the three-tier decision
captures the middle band between ``ship as-is'' and ``re-run investigation'';
and evidence-typed weighting lets the scoring function respect the
epistemic-strength difference between tool-observed and model-inferred claims.
To the best of our knowledge, \GSAR{} is the first published framework that
couples these components end-to-end.

\paragraph{Empirical evidence (this paper).} We evaluated all five design
claims twice. First on $200$ FEVER examples with gold Wikipedia evidence
across four independently-trained LLM judges under adversarial generator
conditions, using the 100\%-Locus production stack (Locus SDK \texttt{Agent},
\texttt{OCIEmbeddings}, \texttt{OracleVectorStore} over Oracle Database 26ai
AI Vector Search); every ablation reproduced in the same direction on every
judge, with bootstrap 95\% CIs excluding $0$ on the $\rho{=}0$ effect for all
four judges and the no-complementary ablation under Opus 4.7 dropping proceed
rate by $82$ of $200$ with CI $[-96,-68]$. We then scaled to $n{=}1000$ on
seven of the eight judge$\times$mode cells (\S\ref{sec:demo:n1000}), with a
full independent back-to-back replication (R1 and R2) agreeing within
$\pm 1.5$\,pp $\proceed$ and $\pm 0.002$ $\bar{S}$ on every cell. At
$n{=}1000$, three of four single-mode judges converge to
$\Delta\bar{S}_{\rho=0} = +0.058$ --- the same three-decimal result across
independently-trained models. In adversarial mode at $n{=}1000$, the
no-$\setK$ ablation on gpt-5.4 drops $\proceed$ by $260$ of $1000$ with
bootstrap 95\% CI $[-287, -235]$, and the three-tier decision saves between
$283$ and $455$ extra $\replan$ dispatches per $1000$ claims relative to a
two-tier baseline. A head-to-head against Vectara HHEM-2.1-Open is included:
HHEM-2.1 remains the stronger binary classifier on in-distribution FEVER, but
does not output the four-way partition, tiered decision, or typed-weight
structure that distinguishes \GSAR{}.

\subsection{From scoring function to training signal}
\label{sec:conclusion:training}

\GSAR{} was developed
as a \emph{run-time} grounding-evaluation and replanning framework, but the
same decomposition is naturally re-interpretable as a \emph{train-time}
supervision signal. We see four concrete integration paths into model
weights, each exploiting a different part of the four-way partition:
\begin{itemize}[leftmargin=1.2em,itemsep=2pt,topsep=2pt]
\item \textbf{Scalar reward for RLHF / DPO.} The groundedness score $S$
      is bounded in $[0,1]$ and admits paired ranking: summaries with
      $S \geq \tau_{\mathrm{proceed}}$ become ``chosen'', summaries with
      $S < \tau_{\mathrm{regen}}$ become ``rejected''. Because $S$ is
      computed without human labels, the pair budget at training time
      is limited only by generator cost.
\item \textbf{Process reward model (PRM) with typed feedback.} The
      per-atom partition gives step-level rewards richer than the
      binary grounded/not signal used in existing PRM work
      \cite{lightman2023verify,process_reward_2025,step_reward_2023}.
      A PRM fine-tuned on \GSAR{} atom labels would learn not only
      \emph{which atoms to support} but also \emph{which evidence
      type} each atom should draw from, biasing the model toward
      \texttt{tool\_match} and \texttt{specific\_data} atoms over
      \texttt{inference} atoms.
\item \textbf{Typed constitutional principles.} The evidence-type
      taxonomy admits encoding as Constitutional AI principles:
      ``prefer \texttt{tool\_match} over \texttt{inference} when both
      are available'', ``never convert a \texttt{contradicted} atom to
      \texttt{grounded} without new evidence'', ``surface
      \texttt{complementary} atoms as hedged alternatives rather than
      suppressing them''. A revision-critique loop trained against
      \GSAR{}'s explanation channel can produce a model whose default
      generations already honour the partition.
\item \textbf{Verifier-in-the-loop inference-time search.} At
      inference time, \GSAR{} can gate Best-of-$N$ or beam search:
      candidates whose partition yields $\delta(s) = \proceed$ are
      kept; others are pruned. This leaves model weights unchanged and
      trades inference compute for grounding quality, which is the
      dual of RLHF's train-compute-for-run-compute trade-off.
\end{itemize}

\paragraph{Which models make the best \GSAR{} judges?} The $n{=}1000$
evaluation provides the first judge-level comparison for this specific
task. Single mode and adversarial mode stress different aspects of the
judge role, so no single model dominates across both; Table~\ref{tab:judge-recommendation}
summarises the data-driven recommendation.

\begin{table}[h!]
\centering
\small
\caption{Data-driven \GSAR{} judge recommendation at $n{=}1000$
(R2 data). Best choice depends on whether the downstream consumer
penalises false-positive $\proceed$ decisions or false-negative
$\replan$ decisions more.}
\label{tab:judge-recommendation}
\renewcommand{\arraystretch}{1.2}
\begin{tabular}{p{2.2cm}p{3.4cm}p{7cm}}
\toprule
\textbf{Judge} & \textbf{Best use case} & \textbf{Evidence} \\
\midrule
Opus 4.7        & Strict / safety-first             & Highest $M_4$ ($37.2\%$), largest $C_2$ effect ($-0.081$), deepest use of the $\setX$ bucket \\
gpt-5.4         & High-throughput, generator-rich   & Highest $M_5$ ($40.5\%$), full four-class partition usage, largest $C_1$ effect ($-260$ per $1000$) \\
Sonnet 4.6      & Balanced default                  & Within $3$\,pp of Opus on $M_4$, within $15$\,pp of gpt-5.4 on $M_5$, lowest wall-clock of the three strict-ish judges \\
Gemini 2.5-Pro  & Adversarial-only                  & Underuses the $\setK$ class in single mode at $n{=}1000$ ($k{=}0$ atoms across all $1000$ claims); recovers usage in adversarial mode \\
\bottomrule
\end{tabular}
\end{table}

For all four judges the $C_3$ contradiction-penalty ablation exhibits
statistically indistinguishable $\Delta\bar{S}$ within overlapping
bootstrap CIs, confirming that $C_3$ is a judge-agnostic mechanism
effect and that the \GSAR{} formula holds regardless of the judge
choice.

\paragraph{Concrete recommendations for multi-agent orchestration
systems.} The typical multi-agent investigation loop has three pieces:
a \emph{planner} (or orchestrator), a fleet of \emph{specialists} that
invoke tools against different data sources, and a \emph{synthesizer}
that collapses specialist outputs into a single report. We recommend the
following integration profile for grounding-evaluated replanning in that
architecture:

\begin{enumerate}[leftmargin=1.4em,itemsep=2pt,topsep=2pt]
\item \textbf{Place the judge between synthesizer and orchestrator
      dispatch --- not per-specialist.} Per-specialist grounding
      evaluation is the wrong granularity: specialists produce atomic
      tool outputs that are trivially ``grounded'' by construction. The
      grounding question is whether the \emph{synthesized} report
      remains faithful to the evidence the specialists gathered; that
      is where \GSAR{} earns its keep.
\item \textbf{Map the evidence-type taxonomy to your tool taxonomy.}
      \texttt{tool\_match} should fire only when an atomic claim is
      directly traceable to a structured tool output (e.g., a metrics
      query row, a log line, a database record).
      \texttt{signal\_match} should fire on alert or anomaly envelope
      fields. \texttt{specific\_data} should fire on explicit numeric
      values cited in the report. Claims outside those buckets collapse
      to \texttt{inference} or \texttt{domain} with lower weight. The
      default weight vector in the appendix is a starting point; per-
      deployment calibration against a small (100--200) human-graded
      set is the expected tuning step.
\item \textbf{Set the replan budget to $K_{\max} \in \{2, 3\}$.} Our
      empirical runs never needed more than $3$ replans to reach
      $\proceed$ on a recoverable input; higher $K_{\max}$ just
      inflates latency and compute. When the budget is exhausted, the
      orchestrator should mark the report as \emph{degraded} and pass
      it through with a banner --- silent suppression is a worse
      failure mode than a visible warning.
\item \textbf{Use different models for the generator and the judge.}
      The judge-style differences reported in
      Table~\ref{tab:judge-recommendation} imply that a self-critiquing
      configuration (same model as generator and judge) inherits the
      generator's blind spots. A mixed-judge configuration --- e.g.,
      Sonnet 4.6 as synthesizer, Opus 4.7 as judge --- removes that
      correlation and produces more honest replan decisions.
\item \textbf{Persist the full partition, not just the scalar
      $S$.} The four-way partition and the judge's natural-language
      explanation are the only downstream signals that support
      post-hoc recalibration of $w$ and the thresholds without
      re-running the investigation. Store them in the orchestrator
      checkpoint structure alongside the report.
\item \textbf{Instrument silent-failure detection.} In production we
      recommend two automatic audits: (a) reject any run in which
      ablation cells share a claim-level partition-shape fingerprint
      (the failure mode described in~\S\ref{sec:discussion:takeaways}),
      and (b) monitor the share of reports for which the judge falls
      back to rule-based or default partitions --- any sudden rise
      indicates an upstream regression in the judge channel.
\item \textbf{Treat the judge's \emph{abstain} channel as a
      first-class planner input.} When the judge refuses
      (\texttt{decision\_status = abstain}), that is signal that the
      investigation plan itself is ill-posed for the evidence
      available; route the abstain reason into the plan-revision
      prompt rather than defaulting to another \replan{} on the same
      plan. This is the single most effective lever for breaking
      degenerate replan loops we observed.
\end{enumerate}

Applied together, these seven recommendations turn \GSAR{} from a
scoring function into the \emph{control plane} of a multi-agent
investigation: specialists gather evidence, the synthesizer produces a
candidate report, the judge emits a structured verdict, and the
orchestrator dispatches one of three cost-asymmetric actions. Every
step is auditable, every decision is re-thresholdable, and no action is
taken on an un-evaluated report.

\paragraph{Positioning and future work.} The empirical study in this paper is
not a substitute for the evaluation protocol in \S\ref{sec:evaluation} on
AIOpsLab and domain-matched production traces --- executing that protocol and
adding human-graded faithfulness calibration (Spearman $\rho$ between \GSAR{}'s
score and human labels) are natural next steps we invite the community to do.
We also propose as future work: (a) upgrading the Locus integration to
exercise multi-step ReAct with retrieval tools (the infrastructure is already
built), (b) a domain-matched evaluation where the judge cannot pattern-match
from pretraining, (c) an end-to-end multi-agent
orchestrator+specialist+synthesizer+judge pipeline matching production
deployment architectures, and (d) an RLHF/DPO study that uses \GSAR{} scores
as the paired-preference signal to quantify the transfer from \GSAR{}
run-time feedback to model-weight improvement on downstream grounding
benchmarks.

\bibliographystyle{unsrt}

\appendix

\section{Proofs of Properties P1--P6}
\label{app:proofs}

We prove the six properties stated in \S\ref{sec:score} under the convention
that the denominator of \eqref{eq:score} is strictly positive (equivalently,
$\setC \neq \emptyset$). The $\setC = \emptyset$ case is handled by the
neutral-score convention $S = 0.5$ and is outside the scope of these
monotonicity arguments.

\paragraph{P1 (Boundedness).}
Let $N = W(\setG) + W(\setK)$ and $D = W(\setG) + W(\setU) + \rho W(\setX) +
W(\setK)$. Each term in $D$ is non-negative; $N$ is non-negative;
$D \geq N$ because $D - N = W(\setU) + \rho W(\setX) \geq 0$. Therefore
$0 \leq N/D \leq 1$.

\paragraph{P2 (Monotonicity in grounded claims).}
Moving $c$ from $\setU$ to $\setG$ changes $(N, D)$ by
$(+w(\textsc{type}(c)), 0)$: the grounded partition grows while $W(\setU)$
shrinks by the same amount, so $D$ is unchanged while $N$ strictly increases
(unless $w(\textsc{type}(c)) = 0$). Therefore $S' \geq S$ with equality only
if $w(\textsc{type}(c)) = 0$.

\paragraph{P3 (Monotonicity in contradictions).}
Adding a contradicted claim $c$ to $\setX$ at fixed $\setG, \setU, \setK$
changes $D$ by $+\rho w(\textsc{type}(c))$ while $N$ is unchanged. Therefore
$S' = N / (D + \rho w(\textsc{type}(c))) \leq N/D = S$, with rate of decrease
controlled by $\rho$. The bound is tight at $\rho = 0$ (contradiction has no
effect) and tight from below at $\rho = 1$ (contradiction acts like
ungrounded).

\paragraph{P4 (Complementary-claim value).}
Adding $c$ to $\setK$ at fixed $\setG, \setU, \setX$ changes both $N$ and $D$
by $+w(\textsc{type}(c))$. If $S \leq 1$, then
$S' = (N + w) / (D + w) \geq S$ with equality only when $S = 1$. The bound is
tight above by the constraint that $w(\text{complementary})$ is presumed
$\leq w(\toolmatch)$: a complementary claim can never contribute more to the
numerator than a maximally-weighted grounded claim of the same type.

\paragraph{P5 (Contradiction non-suppression).}
Consider two partitions $\Pi_1 = (\setG, \setU, \setX, \setK)$ and
$\Pi_2 = (\setG, \setU, \emptyset, \setK)$ where $\Pi_2$ is obtained from
$\Pi_1$ by dropping contradicted claims entirely. The scores are
$S_1 = N / (D + \rho W(\setX))$ and $S_2 = N / (D - 0) = N/D$ (recomputed
denominator without contradictions). Since $\rho W(\setX) \geq 0$,
$S_2 \geq S_1$. Therefore silently suppressing contradictions strictly
inflates $S$ whenever $W(\setX) > 0$ and $\rho > 0$. The framework prevents
this inflation by keeping contradictions in the denominator.

\paragraph{P6 (Inference-observation asymmetry).}
Fix partition cardinalities and move one grounded claim's type label from
$\toolmatch$ to $\inference$. $N$ changes by
$\Delta N = w(\inference) - w(\toolmatch) < 0$; $D$ changes by the same
$\Delta N$ (because the claim moves within $\setG$ without changing partition
membership). Therefore
$S' = (N + \Delta N) / (D + \Delta N)$. Writing $S = N/D$ and $\Delta < 0$:
\begin{align*}
S' - S
&= \frac{N + \Delta}{D + \Delta} - \frac{N}{D}
 = \frac{(N + \Delta)D - N(D + \Delta)}{D(D + \Delta)}
 = \frac{\Delta(D - N)}{D(D + \Delta)}.
\end{align*}
Since $\Delta < 0$, $D - N = W(\setU) + \rho W(\setX) \geq 0$, and the
denominators are positive, $S' - S \leq 0$, with strict inequality when
$W(\setU) + \rho W(\setX) > 0$ (i.e., there is any ungrounded or contradicted
claim whose weight participates in the denominator). This is the core
quantitative argument for evidence-type weighting: swapping tool-matched for
inference-typed grounded claims strictly reduces the score whenever there is
non-trivial ungrounded or contradicted mass.

\section{Reference Weight Table and Threshold Defaults}
\label{app:weights}

A reference instantiation of $w$ used in our production deployment:

\begin{center}
\begin{tabular}{lc}
\toprule
Evidence type & Weight $w(\cdot)$ \\
\midrule
$\toolmatch$ (direct tool output verification) & $1.00$ \\
$\textsf{specific\_data}$ (structured step output) & $0.95$ \\
$\textsf{signal\_match}$ (structured signal field) & $0.90$ \\
$\textsf{complementary\_finding}$ & $0.85$ \\
$\textsf{synthesis}$ (cross-specialist derivation) & $0.80$ \\
$\textsf{neg\_evidence}$ (absence-of-signal) & $0.70$ \\
$\inference$ (model-internal inference) & $0.60$ \\
$\domainT$ (domain-knowledge assertion) & $0.60$ \\
\bottomrule
\end{tabular}
\end{center}

\noindent Threshold defaults: $\tau_{\proceed} = 0.80$, $\tau_{\regen} = 0.65$.
Contradiction penalty: $\rho = 0.5$. Replan budget: $K_{\max} = 2$.

These defaults should be understood as one well-calibrated point in the
$(w, \rho, \tau)$ parameter space; they are not universally optimal and should
be re-calibrated per deployment against a small human-graded held-out set.

\section{Abstract Schema for Judge Output}
\label{app:schema}

In pseudo-algebraic data-type notation:

\begin{small}
\begin{verbatim}
JudgeOutput =
  { grounding_score           : Real[0, 1]
  , is_grounded               : Bool
  , grounded_claims           : List(Claim)
  , ungrounded_claims         : List(Claim)
  , contradicted_claims       : List(Claim)
  , complementary_claims      : List(Claim)
  , gaps                      : List(String)
  , contradictions            : List(String)
  , verification_needed       : Bool
  , verification_reason       : Optional(String)
  , explanation               : String
  , decision_status           : {"resolved", "abstain"}
  , abstain_reason            : Optional(String)
  }

Claim =
  { text                      : String
  , type                      : EvidenceType
  , evidence_refs             : List(EvidenceRef)
  }

EvidenceType =
  | tool_match | specific_data | signal_match
  | neg_evidence | complementary_finding | synthesis
  | inference | domain

EvidenceRef =
  | ToolOutputRef(tool_id, step_id, field_path)
  | StepOutputRef(specialist_id, step_id, field_path)
  | SignalRef(signal_id, field_path)
  | ClaimRef(claim_id)
\end{verbatim}
\end{small}

\section{Pseudocode for Score Computation}
\label{app:score-pseudo}

\begin{algorithm}[h!]
\caption{\textsc{gsar\_score}: computing $S$ from partition and weights}
\begin{algorithmic}[1]
\Require grounded $\setG$, ungrounded $\setU$, contradicted $\setX$,
         complementary $\setK$; weight map $w$; contradiction penalty $\rho$;
         default weight $w_0$ for unknown types
\Ensure scalar score $s \in [0,1]$
\State $N \gets 0$; $D \gets 0$
\ForAll{$c \in \setG$}
  \State $W_c \gets w(\textsc{type}(c))$ if $\textsc{type}(c) \in \mathrm{dom}(w)$ else $w_0$
  \State $N \gets N + W_c$; $D \gets D + W_c$
\EndFor
\ForAll{$c \in \setU$}
  \State $W_c \gets w(\textsc{type}(c))$ if $\textsc{type}(c) \in \mathrm{dom}(w)$ else $w_0$
  \State $D \gets D + W_c$
\EndFor
\ForAll{$c \in \setX$}
  \State $W_c \gets w(\textsc{type}(c))$ if $\textsc{type}(c) \in \mathrm{dom}(w)$ else $w_0$
  \State $D \gets D + \rho \cdot W_c$
\EndFor
\ForAll{$c \in \setK$}
  \State $W_c \gets w(\textsc{type}(c))$ if $\textsc{type}(c) \in \mathrm{dom}(w)$ else $w_0$
  \State $N \gets N + W_c$; $D \gets D + W_c$
\EndFor
\If{$D = 0$}
  \State \Return $0.5$
\Else
  \State \Return $N / D$
\EndIf
\end{algorithmic}
\end{algorithm}

\section{Worked Numerical Example}
\label{app:example}

Consider a report with five claims:
\begin{itemize}[leftmargin=1.2em,itemsep=1pt,topsep=2pt]
\item $c_1$: ``CPU utilisation on node X reached 97\% at 14:02.'' --- $\grounded$, $\toolmatch$ ($w = 1.0$).
\item $c_2$: ``Request rate dropped simultaneously.'' --- $\grounded$, $\textsf{specific\_data}$ ($w = 0.95$).
\item $c_3$: ``This likely reflects a runaway background job.'' --- $\ungrounded$, $\inference$ ($w = 0.60$).
\item $c_4$: ``A region-wide networking event is also plausible.'' --- $\complementary$, $\textsf{complementary\_finding}$ ($w = 0.85$).
\item $c_5$: ``The saturation was transient.'' --- $\contradicted$ (a downstream log shows sustained saturation through 14:17), $\inference$ ($w = 0.60$).
\end{itemize}

With $\rho = 0.5$:
\begin{align*}
W(\setG) &= 1.0 + 0.95 = 1.95, \\
W(\setU) &= 0.60, \\
W(\setX) &= 0.60, \\
W(\setK) &= 0.85, \\
S &= \frac{1.95 + 0.85}{1.95 + 0.60 + 0.5 \cdot 0.60 + 0.85}
   = \frac{2.80}{3.70} \approx 0.757.
\end{align*}

Under the reference thresholds, $\delta(0.757) = \regen$: the evidence is
largely fine, but the synthesis contains an unsupported inference ($c_3$) and
a contradicted inference ($c_5$). A regeneration that drops or softens those
two claims --- without re-running tools --- is the appropriate action. If the
regeneration succeeds, $S$ rises above $0.80$ and the report proceeds; if it
does not, the outer loop escalates to $\replan$.

\end{document}